\documentclass{article} 
\usepackage{iclr2027_conference,times}

\usepackage{amsmath,amsfonts,bm}

\def\eqref#1{equation~\ref{#1}}

\def\1{\bm{1}}

\DeclareMathAlphabet{\mathsfit}{\encodingdefault}{\sfdefault}{m}{sl}
\SetMathAlphabet{\mathsfit}{bold}{\encodingdefault}{\sfdefault}{bx}{n}

\usepackage{hyperref}
\usepackage{url}
\usepackage{graphicx}
\usepackage{subcaption}
\usepackage{booktabs} 
\usepackage{amssymb}
\usepackage{comment}

\usepackage{xcolor}
\newcommand{\revised}[1]{\textcolor{black}{#1}}

\title{SSF-Net: A Spatial–Spectral Fusion Network for Accurate PDE Operator Learning}

\author{anonymous}

\iclrfinalcopy 

\author{
Siqi Zhao \\
University of New South Wales \\
\texttt{siqi.zhao4@student.unsw.edu.au}
\And
Flora D. Salim\thanks{Corresponding author.} \\
University of New South Wales \\
\texttt{flora.salim@unsw.edu.au}
}

\begin{document}

\maketitle

\begin{abstract}
Machine learning methods for solving partial differential equations (PDEs) significantly improve computational efficiency over traditional numerical solvers such as finite element method (FEM). However, PDE dynamics inherently span both local and global dynamics: disturbances propagate and decay gradually through the domain, while global coupling effects can influence the whole system simultaneously. Capturing both types of dynamics with a low complexity model remains a key challenge. To address this, we propose the Spatial-Spectral Fusion Network (SSF-Net), built around a dual-branch Spatial–Spectral Fusion (SSF) block comprising a spatial branch and a spectral branch.  The spatial branch employs a multi-directional Mamba, which mitigates its inherent causal, single-direction scanning bias and capture local interactions with reduced directional bias at linear complexity. The spectral branch leverages a Fourier transform with a block-diagonal MLP to efficiently capture global spectral features with reduced parameter count and complexity. The SSF block is embedded in a multiresolution convolutional encoder-decoder, which extracts multi-scale features. We evaluate SSF-Net on a comprehensive PDE benchmark spanning multiple task types and assess its long-horizon stability. Compared with attention-based and convolution-based baselines, SSF-Net achieves lower prediction error with fewer parameters while retaining near-linear asymptotic complexity in the number of spatial tokens.
\end{abstract}

\section{INTRODUCTION}

Scientific and engineering problems such as aerodynamics and wave propagation involve solving complex partial differential equations (PDEs). Conventional PDE solvers such as the finite element method (FEM), the finite volume method (FVM), and spectral methods discretize the continuous physical domain into a finite set of points, cells, or elements \citep{traditionalpde}. These methods are accurate and physically grounded, but remain computationally expensive at scale. This challenge has motivated the development of data-driven machine learning methods for solving PDEs, among which neural operators have emerged as a particularly promising direction.

Neural operators aim to learn the underlying PDE solution operator, which directly maps an input function space—including initial and boundary conditions, coefficients, etc.—to the corresponding PDE solution \citep{neuraloperator}. Compared to conventional simulation, these methods achieve lower computational cost and faster inference \citep{pdebench}. Moreover, because neural operators are data-driven, they remain applicable even when the governing equations are incomplete or unknown. Existing neural operators are built on a variety of backbones, including convolutions \citep{cno}, graph neural networks \citep{gno}, Fourier transforms \citep{fno,geo-fno,gino}, and transformers \citep{transolver,transolver++,poseidon,pde-transformer}.

Nevertheless, PDE dynamics inherently span both local and global scales. Locally, disturbances such as shock fronts, boundary layers, and localized stress concentrations propagate gradually, decay with distance, and are primarily governed by neighboring interactions. Globally, coupling effects—such as pressure fields in incompressible flow or potential fields in Poisson-type equations—can instantaneously influence the entire system, independent of spatial distance.

\begin{figure}[t]
    \centering
    \includegraphics[width=\linewidth]{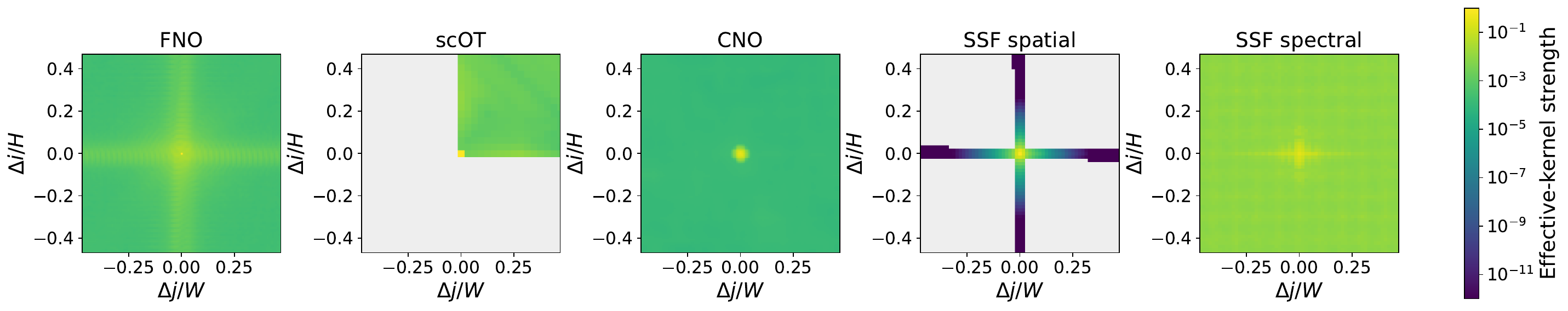}
    \caption{Effective-kernel strength on FNS-KF. Effective-kernel strength is a measure of how strongly infinitesimal changes in the source features affect the query output. See Section~\ref{sec:model_analysis} and Appendix~\ref{app:model_analysis} for details. Axes denote normalized spatial displacements. All panels share a logarithmic color scale, with gray indicating zero.}
    \label{fig:effective-kernel}
\end{figure}

Existing methods struggle to efficiently integrate local and global dynamics. Firstly, convolution- and graph-based models typically require multiple layers to capture long-range dependencies. Fourier-based models can lose fine-scale spatial details through mode truncation. Transformer-based models face a trade-off between the quadratic cost of global attention and the restricted receptive fields of window-based attention. Moreover, these methods do not explicitly model decaying local interaction. Figure~\ref{fig:effective-kernel} illustrates how input locations across the spatial domain influence the output at a given query position. CNO exhibits domain-wide responses concentrated near the query, while FNO shows broad spatial coverage with enhanced influence along the horizontal and vertical axes through the query. The scOT block is restricted to its attention window. None of them shows an explicit gradual decaying influence with distance which is common in the local interaction of PDEs.

To address these gaps, we introduce SSF-Net, a dual-branch neural operator that jointly models local and global PDE dynamics at near-linear complexity. The spectral branch extracts global dynamics via a Fourier transform, while the spatial branch extracts local dynamics via a state-space model (SSM). This design combines complementary local and global representations and achieves improved prediction accuracy with fewer parameters than the compared baselines. As shown in Figure~\ref{fig:effective-kernel}, the SSF spatial branch exhibits a concentrated, directional response that rapidly weakens away from the query, while its spectral branch provides broad spatial coverage. These complementary patterns support the integration of local sensitivity and global interaction within PDEs. 
We make the following contributions:

\begin{itemize}
    \item We propose a novel dual-branch architecture that combines SSM \citep{mamba,lamo} for local dynamic extraction with a Fourier transform for global dynamic extraction. The resulting model, termed the Spatial-Spectral Fusion Network (SSF-Net), is embedded within a U-Net-style multiresolution encoder-decoder to extract multi-scale features while reducing parameter count.

    \item We show that SSF-Net has near-linear asymptotic complexity in the number of spatial tokens, avoiding the quadratic token-mixing cost of full self-attention.

    \item We evaluate SSF-Net on a diverse collection of PDE benchmarks and show that it consistently achieves lower prediction errors with fewer parameters than the compared baselines.

\end{itemize}

\section{RELATED WORK}
Neural operators learn mappings between function spaces to approximate PDE solution operators \citep{neuraloperator}. \citet{operator-categories} classify the neural operators into three categories: deep operator networks (DeepONets), integral kernel operators, and transformer-based neural operators. 

DeepONet combines branch-network features of the input function with trunk-network features of query coordinates through an inner product to evaluate nonlinear operators at arbitrary locations \citep{DeepONet}. Its original architecture does not explicitly separate local and global interactions.

Integral kernel operators approximate the PDE solution function via parameterized kernels. FNO and its variants \citep{fno,AdaptiveFno,drift-net} implement the kernel operator in the Fourier domain. These methods model global interactions through Fourier-domain processing. Geo-FNO \citep{geo-fno}, GINO \citep{gino} and U-FNO \citep{u-fno} generalize the method to irregular geometries and multiphase flow. In contrast, GNO \citep{gno} and CNO \citep{cno} instead parameterize the kernel using GNNs or CNNs, respectively, to model local interactions through spatial neighborhood aggregation.

Transformer-based neural operators discretize the continuous domain, treating each grid point as a token, and use attention as the kernel \citep{transolver,transolver++,Aroma,Cvit}. Transformer-based operators are also widely used in multi-task foundation models\citep{poseidon,pde-transformer,P3D,multiple}. Full self-attention permits direct interactions across the domain through all-to-all token aggregation. However, the quadratic complexity limits its training efficiency. Window-based attention, such as SWIN \citep{swin,poseidon,pde-transformer}, mitigates this cost by restricting attention to local windows, achieving linear complexity. Interactions beyond an individual window generally require propagation through shifted windows, stacked layers, or the multiscale hierarchy.

Beyond this taxonomy, SSMs have recently emerged as a fourth paradigm for neural operators, leveraging recurrent state updates to achieve linear-complexity sequence modeling \citep{Alias-freeMamba,lamo}. Although SSMs can model long-range interactions through recurrent state propagation, distant positions communicate through intervening state updates. Contractive transitions introduce a scan-distance-dependent attenuation bias, while Mamba's selective mechanism makes interactions input-adaptive. 

These architectures provide different mechanisms for spatial interaction, with distinct receptive-field, adaptivity, and computational properties. This motivates SSF-Net, which explicitly fuses spatial and spectral branches to jointly model local and global interactions at near-linear complexity.

\section{METHOD}
\label{sec:method}

\begin{figure}[t]
    \centering
    \includegraphics[width=\linewidth]{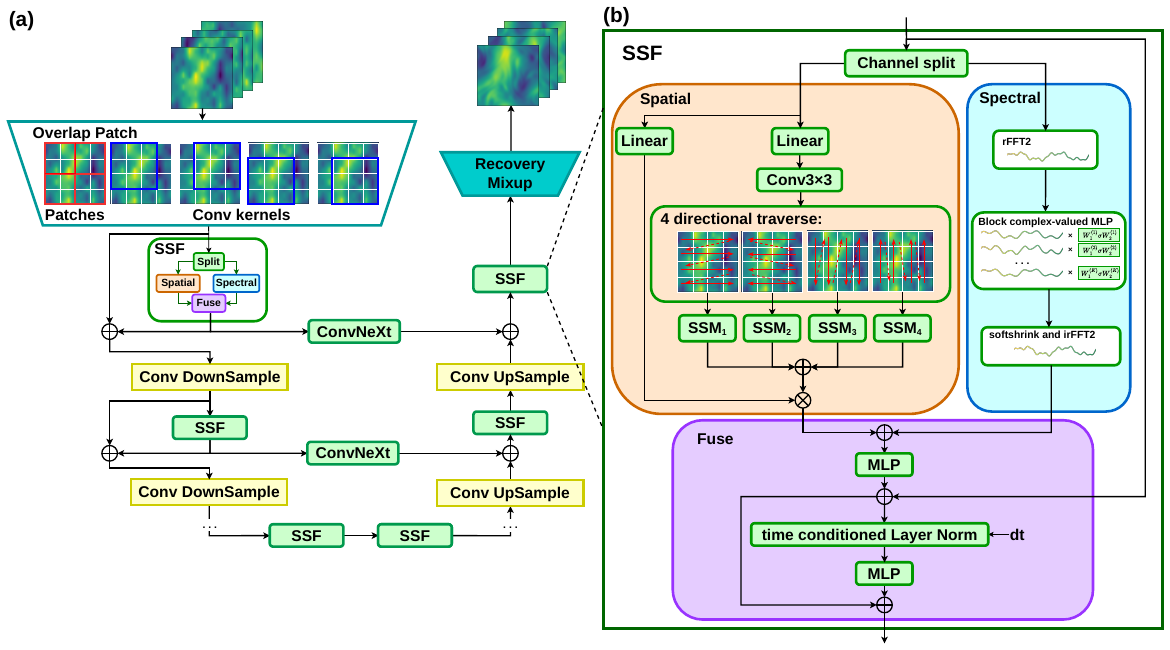}
    \caption{Overview of SSF-Net. (a) U-Net style encoder--decoder with ConvNeXt blocks and overlapping patch embedding. (b) Two parallel branches: a spatial branch with four-directional Mamba scans, and a spectral branch with a rFFT2, a block-diagonal MLP, and soft-shrinkage.}
    \label{fig:overall}
\end{figure}

We propose SSF-Net to jointly model local and global interactions in PDE solution operators with near-linear asymptotic complexity. For time-dependent problems, predictions are generated through autoregressive rollout. The formal problem setting and notation are provided in Appendix~\ref{app:problem_setting}.

\subsection{OVERALL ARCHITECTURE}

As shown in Figure~\ref{fig:overall}(a), SSF-Net adopts a multiresolution encoder-decoder structure, with a dual-branch Spatial-Spectral Fusion (SSF) block introduced at each resolution scale. Within each SSF block, the input is split into two parts, processed independently by the spatial and spectral branches, and then merged (Figure~\ref{fig:overall}(b)). The SSM-based spatial branch emphasizes local interactions, while the multi-directional traverse mitigates the inherent causal bias of SSMs arising from their sequential processing of the input. The FFT-based spectral branch transforms the 2D input into the spectral domain, where nonlinearity is introduced to enhance the model's expressiveness, and a block-diagonal MLP reduces computational cost. Downsampling and upsampling across resolutions are implemented via convolution, and the overlapping patching preserves local continuity across neighboring patches by allowing adjacent patches to share spatial information.

SSF offers the following advantages. (1) The split-process-fuse structure avoids channel inflation, thereby reducing computational cost. (2) At each scale, the spectral branch in SSF achieves a global receptive field without relying on deep stacking or large kernels, as required by convolution-based models \citep{cno}, or long-range sequential propagation, as required by SSM-based models \citep{lamo,mamba}. This enables direct global interactions at each resolution, alongside input-adaptive local interactions in the spatial branch.

\subsection{OVERLAPPING PATCH EMBEDDING}
Neighboring points in PDE data exhibit strong spatial correlations. Instead of partitioning the feature map into disjoint patches, overlapping patch embedding employs a convolutional kernel larger than the stride, allowing neighboring patches to share overlapping receptive fields \citep{PVTv2}. This design facilitates local interactions, preserves spatial continuity across patch boundaries, and alleviates information loss during patching. Specifically, adjacent patch windows are enlarged to overlap by o pixels. Given an input $X_{\mathrm{in}} \in \mathbb{R}^{H \times W \times C}$ and a target patch size $p$, we apply a convolution with kernel size $p+o$, stride $p$ and $C'$ output channels, producing
\begin{equation}
X_{\mathrm{patch}}
= \operatorname{LN}(\operatorname{Conv}(X_{\mathrm{in}})),
\qquad
X_{\mathrm{patch}}
\in \mathbb{R}^{(H/p) \times (W/p) \times C'},
\label{eq:patch_embed}
\end{equation}
where $\operatorname{LN}$ denotes the time conditioned layer normalization defined in \citet{poseidon}. See Appendix \ref{app:patch} for details of overlapping patch, and Appendix \ref{app:depatch} for depatch.

\subsection{SPATIAL-SPECTRAL FUSION (SSF)}
For notational simplicity, we reuse $H, W, C$ to denote the spatial and channel dimensions of the input at the current resolution stage throughout Section~\ref{sec:method}. Their concrete values differ across resolution levels and are updated after each downsampling or upsampling operation. 

\paragraph{Feature split and fusion.}
To avoid channel inflation caused by the dual-branch structure, the input $X \in \mathbb{R}^{H \times W \times C}$ is split along the channel dimension into two parts, processed separately by the spatial and spectral branches. The outputs of the two branches are then fused (Figure~\ref{fig:overall}(b)):
\begin{equation}
X_s, X_f = \operatorname{split}(X),
\qquad
X_s, X_f \in \mathbb{R}^{H \times W \times D},
\label{eq:split}
\end{equation}
\begin{equation}
X_{\mathrm{local}} = f_{\mathrm{spatial}}(X_s),
\qquad
X_{\mathrm{global}} = f_{\mathrm{spectral}}(X_f),
\label{eq:branches}
\end{equation}
\begin{equation}
X_{\mathrm{out}}
= \operatorname{MLP}\bigl(
    \operatorname{LN}(X_{\mathrm{local}} + X_{\mathrm{global}})
  \bigr),
\label{eq:fusion}
\end{equation}
where $D = C/2$. $f_{\mathrm{spatial}}$ and $f_{\mathrm{spectral}}$ denote the spatial and spectral branches, detailed in the following paragraphs. $\operatorname{LN}$ denotes the time conditioned layer normalization defined in \citet{poseidon} over the channel dimension and $\operatorname{MLP}: \mathbb{R}^{D} \to \mathbb{R}^{C}$ acts pointwise, lifting the fused features back to the block width. $X_{\mathrm{out}}$ is the output of the SSF block, which is combined with a residual connection before being passed to the downsampling/upsampling operator. See Appendix \ref{app:SSF-mlp} for details of MLP.

\paragraph{Multi-directional Mamba.} We now detail the spatial branch $f_{\mathrm{spatial}}$ introduced in \eqref{eq:branches}. Following the scan patterns in \citet{dim} and \citet{liu2024vmamba}, we employ a four-directional Mamba traversal as our spatial branch to emphasize local interactions. SSMs are originally designed for sequential data with a causal structure, where each step depends only on the preceding tokens. Spatial interactions in PDE fields need not follow a causal scan ordering and can occur in all directions. To bridge this gap, we introduce a multi-directional traverse mechanism.
As shown in the spatial branch in Figure~\ref{fig:overall}(b), the input is scanned along four paths: top-left to bottom-right, bottom-right to top-left, top-right to bottom-left, and bottom-left to top-right. Each traversal follows a raster ordering (each row or column is scanned in a consistent direction). Each path defines an ordering $\pi_n$ that flattens the 2D grid into a 1D sequence:
\begin{equation}
X_n = \pi_n\bigl(
    \operatorname{Conv}(\operatorname{Linear}(X_s))
  \bigr),
\qquad n = 1,2,3,4,
\label{eq:traverse}
\end{equation}
where the linear projects $X_s \in \mathbb{R}^{N\times D}, \; N = H\times W$ to $X_n \in \mathbb{R}^{N\times D_E}, \; D_E=E \cdot D$, where $E$ is the expansion factor in Mamba. The four sequences contain identical elements and differ only in their ordering. See Appendix \ref{app:spatial_branch} for details of multi-directional traverse.

To better exploit the four traversal sequences, we adopt a multi-head SSM similar to multi-head attention \citep{attention}. The channel dimension $D_E$ of each sequence is split into $M$ heads, each of dimensionality $D_E/M$ stacking across the four traversal directions yields $4M$ heads in total. Within each direction, the outputs of all heads are concatenated, and the resulting representations from the four directions are summed:
\begin{equation}
X_{\mathrm{ssm}}
= \sum_{n=1}^{4}
  \pi_n^{-1}\Bigl(
    \operatorname{Concat}\bigl(
      \operatorname{SSM}(X_n^1),
      \ldots,
      \operatorname{SSM}(X_n^M)
    \bigr)
  \Bigr),
\label{eq:ssm}
\end{equation}
\begin{equation}
X_{\mathrm{local}}
= \operatorname{Linear}\bigl(
    \sigma(X_{\mathrm{ssm}})
    \otimes
    \sigma(\operatorname{Linear}(X_s))
  \bigr),
\label{eq:localout}
\end{equation}
where $n$ indexes the four traversal directions, and $X_n^h \in \mathbb{R}^{N \times D_E/M}$ denotes the $h$-th head ($h=1,\ldots,M$) of the $n$-th traversal sequence $X_n$. $\pi_n^{-1}$ restores each output to the original grid layout so that the four directions are spatially aligned before summation. We use GeLU \citep{gelu} as the activation function $\sigma$. The output $X_{\mathrm{local}} \in \mathbb{R}^{H \times W \times D}$ is the aggregated spatial feature, passed to the subsequent fusion step. See Appendix~\ref{app:ssm} for underlying state-space formulation and Appendix \ref{app:spatial_branch} for implementation details of Mamba.

\paragraph{Adaptive FNO.} 
Following AFNO \citep{AdaptiveFno}, the spectral branch mixes tokens in the frequency domain. Given the spectral input $X_f$, we first apply a 2D FFT:
\begin{equation}
Z = \operatorname{RFFT2D}(X_f),
\qquad
Z \in \mathbb{C}^{H \times W_r \times D},
\quad
W_r = \lfloor W/2 \rfloor + 1,
\label{eq:fft}
\end{equation}
To reduce the parameter count of channel mixing, the channel dimension is partitioned into $K$ blocks, and a two-layer complex-valued MLP is applied independently within each block and shared across all tokens:
\begin{equation}
\tilde{Z}^{(\ell)} = W_2^{(\ell)}\,\sigma\big(W_1^{(\ell)} Z^{(\ell)} + b_1^{(\ell)}\big) + b_2^{(\ell)}, \qquad \ell = 1, \ldots, K,
\label{eq:blockmlp}
\end{equation}
where $W_1^{(\ell)}, W_2^{(\ell)} \in \mathbb{C}^{D/K \times D/K}$ and $\sigma$ denotes ReLU applied to the real and imaginary parts separately. Since frequency modes are sparse, with most energy concentrated in low frequencies, we sparsify $\tilde{Z}$ via soft-shrinkage:
\begin{equation}
\hat{Z} = S_\lambda(\tilde{Z}),
\qquad
S_\lambda(x)
= \operatorname{sign}(x)\max\{|x|-\lambda,0\},
\label{eq:softshrink}
\end{equation}
where $\lambda$  is a fixed tuning parameter that controls the sparsity. After transforming back to the spatial domain via an inverse FFT, a residual connection preserves the branch input alongside the learned global interactions:
\begin{equation}
X_{\mathrm{global}}
= \operatorname{IRFFT2D}(\hat{Z}) + X_f.
\label{eq:ifft}
\end{equation}
See Appendix \ref{app:spectral_branch} for details of nonlinearity on complex domain.

An operator-learning interpretation of the two branches and their fusion is provided in Appendix~\ref{app:operator_view}.

\subsection{MULTIRESOLUTION ENCODER-DECODER}

To extract multi-scale features and reduce the number of high-resolution spatial tokens processed at deeper stages, SSF-Net progressively downsamples the feature map across multiple resolution levels in the encoder, and symmetrically upsamples it in the decoder (Figure~\ref{fig:overall}(a)). See Appendix \ref{app:multiresolution} for details.

Both downsampling and upsampling are implemented via convolution rather than patch merging \citep{swin}: downsampling uses a strided convolution that halves the spatial resolution while doubling the channel dimension, and upsampling uses a symmetric transposed convolution that doubles the spatial resolution while halving the channel dimension. This convolutional design provides a learnable, spatially continuous aggregation that better preserves local smoothness in PDE fields, in contrast to patch merging, which performs a fixed spatial rearrangement followed by a linear projection.

\subsection{COMPLEXITY ANALYSIS}
We analyze complexity in terms of the spatial token count $N=HW$ and channel dimension $D$.

\paragraph{Spatial branch.}
The four-directional Mamba scans cost $O(NDL)$ \citep{mamba,s4}, where $L$ is the SSM state dimension. Linear projections cost $O(ND^2)$, while depthwise convolution and element-wise multiplication each cost $O(ND)$. The total is $O(N(DL+D^2))$, linear in $N$.

\paragraph{Spectral branch.}
The 2D FFT costs $O(ND\log N)$. Block-diagonal channel mixing reduces the MLP cost from $O(ND^2)$ to $O(ND^2/K)$ \citep{AdaptiveFno}, giving $O(ND\log N+ND^2/K)$ overall, near-linear in $N$.

\paragraph{Overall.}
Both branches scale linearly or near-linearly with $N$. The overlap patching, downsample/upsample convolution and convolutional skip modules all scale linearly in $N$ and do not change the scaling. SSF-Net therefore retains a global receptive field with near-linear complexity, avoiding the $O(N^2)$ token-mixing cost of full self-attention. This does not guarantee faster execution: directional scans, spectral transforms, and multiresolution operations incur implementation- and hardware-dependent overhead. \revised{A comparison of measured inference latency and throughput is provided in Appendix~\ref{app:inference_efficiency}.}
 
\section{Experiments}
We assess SSF-Net on a comprehensive dataset against several neural operator baselines in this section. And we also validate the contribution of major components of our model through ablation study.

\subsection{Implementation Details}
\paragraph{Benchmarks.}
We evaluate SSF-Net on twelve two-dimensional PDE benchmarks. From PDE GYM \citep{poseidon}, we include four incompressible flow tasks: NS-PwC tests the transport of discontinuous vorticity, NS-SL captures shear-layer instability and vortex roll-up, NS-Tracer-PwC examines coupled velocity and passive-tracer transport, and FNS-KF tests multiscale flow dynamics under sustained forcing. We further include Poisson-Gauss, CE-RP, Wave-Gauss, and ACE, covering elliptic problems, compressible flow, wave propagation, and reaction--diffusion dynamics. To assess long-horizon prediction, we use four APEBench-generated datasets \citep{koehler2024apebench}: Kolm-Smooth and Kolm-Turb for forced Kolmogorov flow, and KS-Weak and KS-Strong for Kuramoto--Sivashinsky dynamics, each evaluated over 100 autoregressive steps. Dataset details are provided in Appendix~\ref{app:data}.

\paragraph{Baselines}
We evaluate SSF-Net against following baselines: FNO \citep{fno}, CNO \citep{cno}, PDE-Transformer (MC) \citep{pde-transformer}, scOT \citep{poseidon} and Drift-Net \citep{drift-net}.

\paragraph{Training and metrics.}
We train a one-step operator with the relative $L_1$ loss, using the frequency-weighted loss of \citet{drift-net} for Drift-Net, and evaluate time-dependent tasks through autoregressive rollout. We report three relative $L_1$  metrics: $\mathcal{E}_{\mathrm{step}}$, which averages errors equally across forecast steps, $\mathcal{E}_{\mathrm{full}}$, which measures the total trajectory error relative to the total reference magnitude, and $\mathcal{E}_{\mathrm{final}}$, which evaluates the last forecast step. For multiple quantities of interest, errors are computed separately and averaged with equal weight. Formal definitions are provided in Appendix~\ref{app:evaluation_metrics}.

\subsection{Main Results}
\begin{table*}[t]
    \centering
    \caption{
        Test errors on eight PDE benchmarks.
        For time-dependent (TD) tasks, each entry reports
        $\mathcal{E}_{\mathrm{step}}/
         \mathcal{E}_{\mathrm{full}}/
         \mathcal{E}_{\mathrm{final}}$ in percent.
        For the time-independent (TI) Poisson-Gauss task,
        only a single relative $L_1$ error in percent is reported.
        Lower is better.
        The best and second-best results for each metric
        within each dataset are shown in bold and underlined,
        respectively.
    }
    \label{tab:relative_l1_results}
    \resizebox{\textwidth}{!}{%
    \begin{tabular}{llcccccc}
        \toprule
        Dataset & Type & FNO & CNO & PDE-Transformer
        & scOT & Drift-Net & SSF-Net \\
        \midrule
        Params & -- & 36M & 18M & 33M & 20M & 17M & 11M \\
        \midrule
        Poisson-Gauss & TI
        & 14.37
        & 1.16
        & \underline{0.18}
        & \revised{0.43}
        & \revised{0.42}
        & \textbf{0.16} \\

        NS-PwC & TD
        & 0.45/0.45/0.68
        & 2.25/2.25/4.58
        & \underline{0.31}/\underline{0.31}/\underline{0.42}
        & \revised{0.34/0.34/0.60}
        & \revised{0.50/0.50/0.85}
        & \textbf{0.10}/\textbf{0.10}/\textbf{0.17} \\

        NS-Tracer-PwC & TD
        & 1.00/1.00/1.85
        & 2.55/2.53/6.54
        & 0.69/0.68/1.12
        & \revised{\underline{0.48}/\underline{0.48}/\underline{0.82}}
        & \revised{0.86/0.85/1.85}
        & \textbf{0.18}/\textbf{0.18}/\textbf{0.33} \\

        NS-SL & TD
        & 0.68/0.69/1.24
        & 3.23/3.32/7.11
        & \underline{0.30}/\underline{0.31}/\underline{0.56}
        & \revised{0.66/0.67/1.25}
        & \revised{0.86/0.88/1.67}
        & \textbf{0.23}/\textbf{0.24}/\textbf{0.48} \\

        FNS-KF & TD
        & 0.41/0.41/0.63
        & 2.33/2.34/4.80
        & \underline{0.28}/\underline{0.29}/\underline{0.43}
        & \revised{0.31/0.31/0.52}
        & \revised{0.57/0.57/1.03}
        & \textbf{0.10}/\textbf{0.10}/\textbf{0.16} \\

        CE-RP & TD
        & 6.30/6.21/12.43
        & 24.50/24.14/46.26
        & \underline{3.48}/\underline{3.43}/\underline{7.02}
        & \revised{5.06/4.99/9.97}
        & \revised{7.65/7.55/15.20}
        & \textbf{2.75}/\textbf{2.72}/\textbf{5.71} \\

        Wave-Gauss & TD
        & 16.14/15.75/26.83
        & 16.83/16.63/22.04
        & \underline{4.80}/\underline{4.70}/\underline{7.28}
        & \revised{6.30/6.15/9.76}
        & \revised{5.59/5.46/8.52}
        & \textbf{4.43}/\textbf{4.32}/\textbf{7.10} \\

        ACE & TD
        & 1.30/1.45/2.33
        & 3.01/3.18/3.24
        & 1.20/2.32/3.94
        & \revised{\underline{0.11}/\underline{0.12}/\underline{0.21}}
        & \revised{0.12/0.13/0.25}
        & \textbf{0.02}/\textbf{0.02}/\textbf{0.03} \\
        \bottomrule
    \end{tabular}%
    }
\end{table*}

\begin{table*}[t]
    \centering
    \caption{
        Test errors over 100 autoregressive rollout steps
        on the APEBench-generated benchmarks.
        Each entry reports
        $\mathcal{E}_{\mathrm{step}}/
         \mathcal{E}_{\mathrm{full}}/
         \mathcal{E}_{\mathrm{final}}$ in percent.
        Lower is better.
        The best and second-best results for each metric
        within each dataset are shown in bold and underlined,
        respectively.
    }
    \label{tab:long_rollout_results}
    \begin{tabular}{lcccc}
        \toprule
        Dataset & FNO & CNO & PDE-Transformer & SSF-Net \\
        \midrule
        Params & 36M & 18M & 33M & 11M \\
        \midrule
        Kolm-Smooth
        & \underline{2.23}/\underline{2.21}/\underline{4.63}
        & 11.52/11.49/28.48
        & 2.39/2.38/5.19
        & \textbf{1.16}/\textbf{1.16}/\textbf{2.47} \\

        Kolm-Turb
        & 7.86/7.73/23.29
        & 70.40/70.09/115.16
        & \underline{4.13}/\underline{4.08}/\underline{13.58}
        & \textbf{2.00}/\textbf{1.96}/\textbf{6.98} \\

        KS-Strong
        & \underline{3.93}/\underline{3.92}/12.39
        & 24.98/24.90/67.08
        & 4.09/4.08/\underline{11.83}
        & \textbf{0.52}/\textbf{0.52}/\textbf{1.66} \\

        KS-Weak
        & \underline{1.23}/\underline{1.23}/\underline{2.92}
        & 13.36/13.29/38.76
        & 7.18/7.14/15.15
        & \textbf{0.23}/\textbf{0.23}/\textbf{0.53} \\
        \bottomrule
    \end{tabular}
\end{table*}

\paragraph{Overall performance.}
\revised{As shown in Table~\ref{tab:relative_l1_results}, SSF-Net achieves the best performance on all eight benchmarks. On the seven time-dependent tasks, it consistently obtains the lowest step-averaged, full-rollout, and final-step relative $L_1$ errors.} Compared with the strongest baseline for each task, \revised{SSF-Net reduces $\mathcal{E}_{\mathrm{step}}$, $\mathcal{E}_{\mathrm{full}}$, and $\mathcal{E}_{\mathrm{final}}$ by 46.9\%, 47.2\%, and 43.3\% on average, respectively.}  On Poisson-Gauss, it achieves an error of 0.16\%, compared with 0.18\% for PDE-Transformer. These results are obtained with approximately 11M parameters, fewer than all compared baselines.

\paragraph{Performance across PDE families.}
SSF-Net performs particularly well on fluid-dynamical, coupled transport, and reaction--diffusion problems. Relative to the strongest baseline on each task, its step-averaged error is reduced by 67.7\% on NS-PwC, \revised{62.5\% on NS-Tracer-PwC}, 64.3\% on FNS-KF, and \revised{81.8\% on ACE}. The gains are more moderate on NS-SL and CE-RP, with step-averaged error reductions of 23.3\% and 21.0\%, respectively. Improvements are smallest on Wave-Gauss, where SSF-Net nevertheless achieves lower errors under all three metrics. These results suggest that the proposed spatial--spectral fusion is effective across elliptic, parabolic, hyperbolic, fluid-dynamical, and coupled transport problems rather than being specialized to a single equation family.

\begin{figure}[t]
    \centering
    \includegraphics[width=\linewidth]{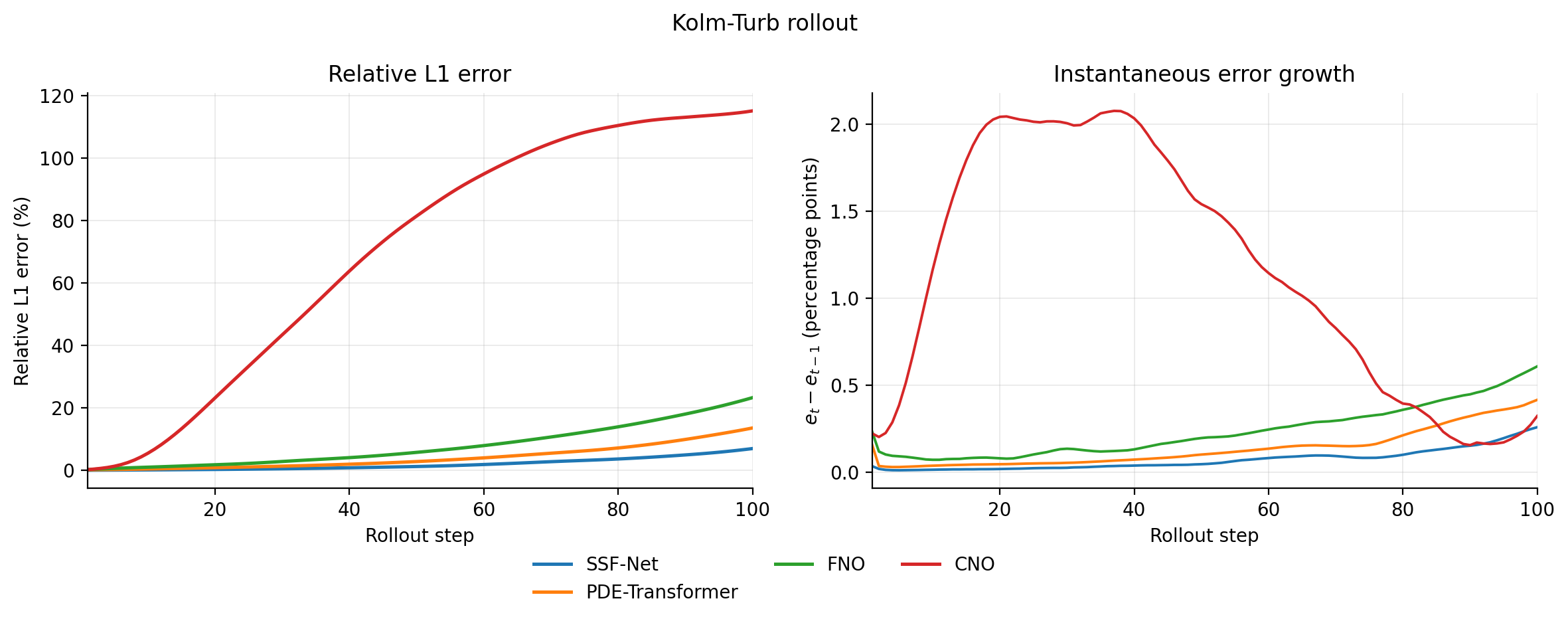}
    \caption{
    Error accumulation over 100 autoregressive rollout steps on Kolm-Turb.
    (a) relative $L_1$ error averaged over test trajectories at each step, reported in percent.
    (b) the corresponding step-to-step error increment $e_t-e_{t-1}$, reported in percentage points, with $e_0=0$.
    }
    \label{fig:kolm-turb-longrollout}
\end{figure}

\paragraph{Long-horizon rollout stability.}
Table~\ref{tab:long_rollout_results} reports a separate 100-step autoregressive evaluation on four APEBench-generated benchmarks, comparing SSF-Net with FNO, CNO, and PDE-Transformer. SSF-Net achieves the lowest errors under all three metrics on all four tasks. Relative to the strongest reported baseline for each task and metric, \revised{it reduces $\mathcal{E}_{\mathrm{step}}$, $\mathcal{E}_{\mathrm{full}}$, and $\mathcal{E}_{\mathrm{final}}$ by 66.9\%, 66.9\%, and 65.8\%, respectively, averaged equally over the four tasks}. Figure~\ref{fig:kolm-turb-longrollout} further examines error accumulation on Kolm-Turb. SSF-Net achieves the lowest test-averaged relative $L_1$ error among the reported methods at every forecast step, reaching 6.98\% at step 100, compared with 13.58\% for PDE-Transformer, the strongest baseline on this task. The error-increment curves show that SSF-Net generally accumulates error more slowly, with a final-step increment of approximately 0.26 percentage points, compared with 0.41 for PDE-Transformer. A smaller late-stage increment alone, however, does not imply accurate prediction: CNO's increments decline substantially after their early peak while its relative error exceeds 100\%. Together, these results support improved empirical rollout stability relative to the reported baselines over the evaluated horizon, although SSF-Net's prediction error continues to grow.

\subsection{Ablation Study}
\label{sec:ablation}

\begin{table}[t]
    \centering
    \caption{
    Ablation study on ACE and FNS-KF. Each entry reports $\mathcal{E}_{\mathrm{step}}/ \mathcal{E}_{\mathrm{full}}/ \mathcal{E}_{\mathrm{final}}$ in percent, lower is better. The best and second-best results for each metric within each dataset are shown in bold and underlined, respectively, including ties.
    }
    \label{tab:ablation}
    \small
    \setlength{\tabcolsep}{5pt}
    \begin{tabular}{lccc}
        \toprule
        Variant & Params & ACE & FNS-KF \\
        \midrule

        Spectral branch only & 9M & 0.021/0.023/0.033 & 0.115/0.115/0.188 \\

        Spatial branch only & 16M & \textbf{0.012}/\textbf{0.014}/\textbf{0.026} & \underline{0.098}/\underline{0.098}/0.168 \\

        1-directional traversal & 10M & \underline{0.020}/\underline{0.021}/\underline{0.031} & 0.106/0.106/0.177 \\

        8-directional traversal & 13M & 0.022/0.024/0.035 & \textbf{0.096}/\textbf{0.096}/\underline{0.159} \\

        \revised{Patch size 4} & \revised{12M} & \revised{0.047/0.052/0.093} & \revised{0.158/0.158/0.238} \\

        \revised{Non-overlapping patches} & \revised{12M} & \revised{0.058/0.065/0.101} & \revised{0.189/0.189/0.322} \\

        \revised{Patch merging/expanding} & \revised{11M} & \revised{0.025/0.027/0.037} & \revised{0.156/0.157/0.266} \\

        \midrule
        Full SSF-Net & 11M & \underline{0.020}/\underline{0.021}/0.033 & \underline{0.098}/\underline{0.098}/\textbf{0.157} \\
        \bottomrule
    \end{tabular}
\end{table}

We evaluate seven variants of SSF-Net: 
(1) retaining only the spectral branch with doubled branch width.
(2) retaining only the spatial branch with doubled branch width.
(3) replacing the four-directional traversal with a single direction.
(4)  extending the four axial traversals to eight by adding forward and reverse scans along both diagonal families (main-diagonal and anti-diagonal directions). 
\revised{(5) increasing the patch size from 2, used in the full model, to 4, matching that of the patch-based baselines. We keep the model backbone configuration invariant. 
(6) replacing the overlapping patch embedding and reconstruction
with their non-overlapping counterparts 
and (7) replacing convolutional downsampling and upsampling with patch merging and expanding, respectively.
An additional comparison with a ConvNeXt-style spatial branch supporting the contribution of the Mamba-based pathway is given in Appendix~\ref{app:conv_spatial_ablation}.
All variants retain the same number of resolution stages, block depths, and encoder--decoder layout. In each single-branch variant, the retained branch is widened to preserve the original block width. }

Table~\ref{tab:ablation} examines 
branch composition and traversal count on ACE and FNS-KF. 
ACE evolves through local nonlinear reaction and diffusion, making spatially adaptive processing a natural fit. Consistent with this structure, the wider spatial-only variant achieves the lowest errors on ACE, although it requires 16M parameters compared with 11M for the full model. FNS-KF combines advective transport with nonlocal pressure coupling under sustained forcing, motivating joint spatial and spectral processing. On this task, the full model achieves the lowest final-step error (0.157\%), outperforming both single-branch variants.  Four-directional traversal improves FNS-KF over a single direction but offers no gain on ACE. The competitive performance of a single direction on ACE may reflect its isotropic reaction--diffusion dynamics, for which additional scan directions provide limited benefit in these experiments. Eight directions yield only slightly lower average errors on FNS-KF at a higher parameter cost. These results suggest that the benefits of fusion and multi-directional traversal depend on the dynamics, with the full configuration offering a practical balance between accuracy and model size.
\revised{To assess the influence of model capacity, we additionally vary the embedding dimension of the single-branch variants while retaining the remaining architectural and training settings (Appendix~\ref{app:capacity_ablation}).}

\revised{Increasing the patch size from 2 to 4 raises the errors on both ACE and FNS-KF, supporting the benefit of finer tokenization on these tasks. Nevertheless, SSF-Net with patch size 4 still achieves lower errors than all evaluated baselines on both tasks, indicating that its advantage cannot be explained solely by the finer token grid. For a $128\times128$ input, patch size 2 produces four times as many initial tokens as patch size 4. Thus, the finer tokenization also increases computational cost, and the lower parameter count should not be interpreted as a matched computational budget.}

\revised{We further examine the contributions of patch overlap and multiresolution resampling. Replacing the overlapping patch embedding and reconstruction with their non-overlapping counterparts increases the final-step error from 0.033\% to 0.101\% on ACE and from 0.157\% to 0.322\% on FNS-KF. Replacing convolutional downsampling and upsampling with patch merging and expanding increases these errors to 0.037\% and 0.266\%, respectively. Both replacements also increase the step-averaged and full-rollout errors on both tasks. These results indicate that patch overlap and convolutional resampling contribute to the full model's accuracy, alongside the spatial--spectral processing.}

\subsection{Model Analysis}
\label{sec:model_analysis}

\begin{figure}[t]
    \centering
    \includegraphics[width=\linewidth]
    {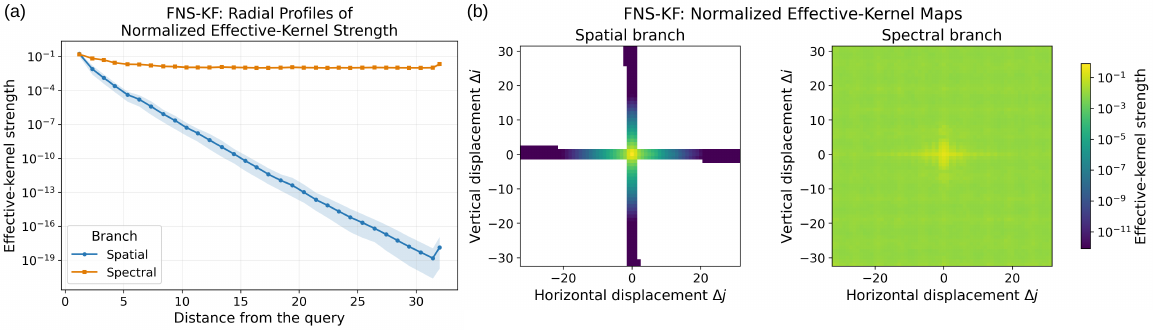}
    \caption{
    Effective-kernel analysis on FNS-KF at the first encoder block. 
    (a) radially averaged normalized kernel strength, showing sample means and 2.5th--97.5th percentile bands. 
    (b) sample-averaged normalized kernel strength of the spatial and spectral branches of SSF-Net. All panels share a logarithmic color scale, with gray indicating zero.
    }
    \label{fig:FNS-KF-effective-kernel}
\end{figure}

To examine whether the two branches primarily focus on different interaction patterns---local dynamics in the spatial branch and global coupling in the spectral branch--we analyze the effective kernels of the first encoder block on FNS-KF using 16 test samples at the initial stored time, with the query fixed at the center of the feature grid. Effective-kernel strength measures how strongly input perturbations at each source location affect the query output, quantified by the Jacobian Frobenius norm across input and output channels. Each kernel is normalized by its full-map $\ell_2$ norm separately for each sample and branch before sample averaging. Implementation details are provided in Appendix~\ref{app:model_analysis}.

As shown in Figure~\ref{fig:FNS-KF-effective-kernel}(b), the spatial branch concentrates its influence near the query in a cross-shaped pattern aligned with the four scan directions, whereas the spectral branch exhibits domain-wide responses with axis-aligned enhancements. Figure~\ref{fig:FNS-KF-effective-kernel}(a) radial profiles quantify this difference: spatial influence decays rapidly and approximately exponentially over most of the measured range, while spectral influence has a much flatter far-field profile. These complementary patterns agree with the analysis in Appendix~\ref{app:operator_view}.

\section{Limitations}
SSF-Net has two main limitations.
First, although its computational complexity is near-linear in the number of spatial tokens, its practical training and inference costs are higher than those of the compared baselines. Multi-directional scans, spectral transforms, and their implementation overhead limit the extent to which favorable asymptotic scaling translates into wall-clock efficiency.

Second, the spatial branch retains direction-dependent effective kernels. Unlike the spectral branch, which enables direct domain-wide interactions, spatial interactions are mediated by scan paths. Consequently, locations that are close in Euclidean distance may remain far apart along the traversal orders, causing their interactions to be strongly attenuated. Multi-directional traversal mitigates this mismatch but does not eliminate it, leaving some nearby interactions weakly represented by the spatial branch.

\section{Conclusion}
We introduced SSF-Net, a neural operator that combines multi-directional Mamba and Fourier-domain processing within a multiresolution encoder--decoder. The two branches provide complementary interaction patterns: spatial responses concentrate near the query and decay rapidly, while spectral responses extend across the domain. Across twelve PDE benchmarks, SSF-Net achieves lower prediction errors with fewer parameters than the compared baselines, and maintains its advantage over the evaluated 100-step autoregressive rollouts. Ablation studies reveal task-dependent trade-offs in branch
composition and traversal count. Although SSF-Net retains near-linear asymptotic complexity, its practical computational overhead and direction-dependent spatial kernels remain limitations. Future work will focus on improving training and inference efficiency and developing spatial interaction mechanisms that better preserve Euclidean neighborhoods. SSF blocks can also replace the windowed attention blocks in \citet{poseidon}, and may be integrated into future foundation models.

\subsection*{AI use statement}
We used generative AI tools to improve the grammar, clarity, and wording of the paper and to assist in checking the correctness of our mathematical derivations. The authors reviewed all AI-assisted suggestions and independently checked the mathematical arguments. We take full responsibility for the final content of the paper.

\bibliography{iclr2027_conference}
\bibliographystyle{iclr2027_conference}

\appendix
\section{PROBLEM SETTING}
\label{app:problem_setting}
We consider PDEs of the general time-dependent form
\begin{equation}
\begin{split}
\partial_t u(x,t) + \mathcal{L}\left(u, \nabla_x u, \nabla_x^2 u, \dots\right) = 0, \quad \forall x \in D \subset \mathbb{R}^d, \quad t \in (0, T), \\
\mathcal{B}(u) = 0, \quad \forall (x,t) \in \partial D \times (0,T), \quad u(0,x) = a(x), \quad x \in D,
\end{split}
\label{pde_de}
\end{equation}
where the solution is defined on a domain $\mathcal{X} \subset L^p(D; \mathbb{R}^n)$ for some $1 \leq p < \infty$ and time horizon $T > 0$. Here $a \in \mathcal{X}$ is the initial condition, and $u \in C([0,T]; \mathcal{X})$ is the corresponding solution. $\mathcal{L}$ and $\mathcal{B}$ denote the differential and boundary operators, respectively. The \eqref{pde_de} also covers PDEs with high-order time derivatives, as well as time-independent PDEs with variable coefficients and source terms.

Time-independent PDEs can be recovered from \eqref{pde_de} by taking the long-time limit $t \to \infty$, i.e., $\bar{u} = \lim_{t\to\infty} u$, yielding the time-independent counterpart
\begin{equation}
\mathcal{L}\left(\bar{u}(x), \nabla_x \bar{u}, \nabla_x^2 \bar{u}, \dots\right) = 0, \quad \forall x \in D, \qquad \mathcal{B}(\bar{u}) = 0, \quad \forall x \in \partial D.
\label{pde_inde}
\end{equation}

The solution of \eqref{pde_de} defines a map $\mathcal{S}: [0,T] \times \mathcal{X} \to \mathcal{X}$, such that for any $t \in [0,T]$ and initial condition $a \in \mathcal{X}$, the solution is $u(t) = \mathcal{S}(t,a)$. For time-independent PDEs, the corresponding solution map is $\bar{\mathcal{S}}: \mathcal{X} \to \mathcal{X}$, with $\bar{u} = \bar{\mathcal{S}}(a)$.

This formulation gives rise to the underlying operator learning task: given a data distribution $\mu \in \mathrm{Prob}(\mathcal{X})$ and an initial condition $a \sim \mu$, our goal is to find an approximation $\mathcal{S}^*$ that closely reproduces the true solution operator $\mathcal{S}$ in \eqref{pde_de}. Given only the initial condition (and boundary conditions), the learned operator should generate the full trajectory $\{\mathcal{S}^*(t,a)\}_{t \in [0,T]}$ that approximates the true trajectory $\{\mathcal{S}(t,a)\}_{t \in [0,T]}$.

While the formulation above is defined over continuous function spaces, in practice we approximate $\mathcal{S}$ using a discretized representation of the domain. We consider a two-dimensional spatial domain $D \subset \mathbb{R}^2$, discretized on a grid of size $H \times W$ with $C$ channels. Our model learns a one-step operator $\mathcal{S}_\theta: u_t \mapsto u_{t+1}$ and is trained with the standard relative $L^1$ loss. At inference stage, given only the initial condition and boundary conditions, it generates the full trajectory $\{u_t\}_{t=1}^{T}$ through autoregressive rollout.

\section{SSM PRELIMINARIES}
\label{app:ssm}
For completeness, we briefly recap the state-space formulation underlying Mamba \citep{mamba}. We refer readers to \citet{s4} for the full derivation. A continuous-time SSM maps an input signal $x(t) \in \mathbb{R}^N$ to an output $y(t) \in \mathbb{R}^N$ through a hidden state $h(t) \in \mathbb{R}^L$. An SSM can be formulated as ordinary differential equations (ODEs) as follows:
\begin{equation}
h'(t) = \mathbf{A}h(t) + \mathbf{B}x(t), \qquad y(t) = \mathbf{C}h(t),
\end{equation}
where $\mathbf{A} \in \mathbb{R}^{L \times L}, \mathbf{B} \in \mathbb{R}^{L \times 1}, \mathbf{C} \in \mathbb{R}^{1 \times L}$ are learnable parameters.

To integrate the SSMs into neural networks, discretization operation is necessary. We use zero-order hold (ZOH) as the discretization rule to discretize SSM with a timescale $\Delta \in \mathbb{R}$:
\begin{equation}
\begin{split}
h[i] = \mathbf{\bar{A}}h[i-1] + \mathbf{\bar{B}}x[i], \qquad y[i] = \mathbf{C}h[i], \\
\mathbf{\bar{A}} = e^{\Delta \mathbf{A}}, \qquad \mathbf{\bar{B}} = (\Delta \mathbf{A})^{-1}(e^{\Delta \mathbf{A}}-\mathbf{I}) \cdot \Delta \mathbf{B},
\end{split}
\end{equation}
\begin{equation}
y = x*\mathbf{\bar{K}},  \qquad
\mathbf{\bar{K}} = (\mathbf{C}\mathbf{\bar{B}},\mathbf{C}\mathbf{\bar{A}}\mathbf{\bar{B}},...,\mathbf{C}\mathbf{\bar{A}}^k \mathbf{\bar{B}},...),
\label{eq:conv_ssm}
\end{equation}
where $\mathbf{\bar{A}}$ and $\mathbf{\bar{B}}$ are the discrete counterparts of $\mathbf{A}$ and $\mathbf{B}$. Equation~\ref{eq:conv_ssm} is the convolution form realization of SSMs.

Mamba further makes $\mathbf{B}$, $\mathbf{C}$, and $\Delta$ input-dependent, allowing the model to selectively propagate or forget information along the sequence. With these input-dependent parameters, the \eqref{eq:conv_ssm} can be written as follows:
\begin{equation}
\begin{split}
\mathbf{B}_t=\mathbf{B}(x_t), \qquad \mathbf{C}_t=\mathbf{C}(x_t), \qquad \Delta_t=\Delta(x_t), \\
\mathbf{\bar{A}}_t = e^{\Delta_t \mathbf{A}}, \qquad \mathbf{\bar{B}}_t = (\Delta_t \mathbf{A})^{-1}(e^{\Delta_t \mathbf{A}}-\mathbf{I})\Delta_t \mathbf{B}_t,
\end{split}
\label{eq:BCdelta_dependent}
\end{equation}
\begin{equation}
y_t=\sum^t_{j=0}\mathbf{K}_{t,j}x_j, \qquad \mathbf{K}_{t,j}=\mathbf{C}_t(\prod^t_{i=j+1}\mathbf{\bar{A}}_i)\mathbf{\bar{B}}_j.
\label{eq:conv_ssm_dependent}
\end{equation}
The matrix of Mamba is a diagonal matrix, $\mathbf{A}=\operatorname{Diag}(\zeta_1,...,\zeta_n)$ with $\zeta_i \in \mathbb{R}, \forall i$. 
Unlike conventional linear time invariance (LTI) SSMs, where the kernel $K$ depends only on the time difference $t - j$, selective SSMs produce an input-dependent kernel whose coefficients vary with both the input sequence and temporal positions. This breaks time-invariance and lets the model decide, per position, how much of the incoming signal to write into the state ($\mathbf{B}$), what to read out of it ($\mathbf{C}$), and how quickly to forget what it already holds ($\Delta$). For PDE fields this is the mechanism by which the branch adapts its effective interaction range to the local state of the solution.

\section{AN OPERATOR-LEARNING VIEW OF SSF}
\label{app:operator_view}
SSF-Net is not discretization-invariant in the sense of
\citet{kovachki2023neural}. It sits with other patch-based surrogates \citep{poseidon,drift-net,pde-transformer}. We nonetheless adopt \eqref{eq:kernel_integral} to characterize the kernels each
branch can represent at a fixed resolution, not to establish a
continuous limit. Both branches of the SSF block can be understood as parameterizations of a single kernel integral operator and differ only in the structure they impose on $\kappa$. 
\begin{equation}
\mathcal{K}(X)(s) = \int_{\Omega} \kappa(s,t)\,X(t)\,\mathrm{d}t, \qquad s \in \Omega,
\label{eq:kernel_integral}
\end{equation}
where $\Omega \subset \mathbb{R}^d$ is the spatial domain on which the discretized field is defined, $s$ and $t$ denote a query and a source location, and $\kappa: \Omega\times\Omega \to \mathbb{R}^{D\times D}$ is a matrix-valued kernel specifying how strongly the feature at $t$ influences that at $s$. This view makes explicit why the two branches are complementary rather than redundant.

\subsection{Spatial branch}
\paragraph{The spatial branch introduces a scan-distance-dependent attenuation.} A discrete selective SSM layer can be interpreted as a Monte-Carlo approximation of the kernel integral \eqref{eq:kernel_integral}, with the kernel parameterized by the state-space dynamics \citep{lamo}:
\begin{equation}
\begin{split}
\kappa_{\mathrm{local}}(s,t)
&= C(s)
   \Bigl(
     \prod_{i=t+1}^{s}\bar{A}(i)
   \Bigr)
   \bar{B}(t), \\
\bar{A}(i)
&= \operatorname{diag}(
     \zeta_{i,1},\zeta_{i,2},\ldots,\zeta_{i,d}
   ),
\end{split}
\label{eq:local_kernel}
\end{equation}
where $\bar{A}(i)$ is diagonal, so the product reduces to an elementwise accumulation along the ordering. Two properties follow. First, the accumulated product acts as an input-dependent relative positional encoding: since $|\zeta_{i,\ell}| \le 1$ for every eigenvalue in a stable SSM, the transition product has a non-increasing magnitude as additional scan steps are accumulated. This introduces a scan-distance-dependent attenuation of $\kappa_{\mathrm{local}}$. Second, $\bar{B}$, $C$ and the $\bar{A}$ are all functions of the input field according to Appendix~\ref{app:ssm}, so $\kappa_{\mathrm{local}}(s,t)$ depends on the two locations separately rather than on their displacement. Consequently, the kernel need not depend only on displacement: both the interaction strength and its rate of attenuation can adapt to the local state of the solution.

\paragraph{The multi-directional traverse symmetrises the local kernel.}
Equation~\ref{eq:local_kernel} describes a single traversal, and \ref{app:spatial_branch} illustrates the multi-directional traverse mechanism. Because the SSM recurrence is causal in
the ordering $\pi_n$, the kernel of direction $n$ vanishes on half of the domain,
\begin{equation}
\kappa_{\mathrm{local}}^{(n)}(s,t) = 0,
\quad \text{whenever} \quad
\pi_n(t) > \pi_n(s),
\end{equation}
writing $\rho = \max_{i,l} |\zeta_{i,l}| \le 1$ for the largest eigenvalue
magnitude and assuming $
\|C(s)\|\le1, \,\|\bar B(t)\|\le1$, its remaining entries obey
\begin{equation}
\bigl\|\kappa_{\mathrm{local}}^{(n)}(s,t)\bigr\|
\lesssim \rho^{\,d_n(s,t)},
\qquad
d_n(s,t) = \bigl|\pi_n(s)-\pi_n(t)\bigr|.
\label{eq:kernel_local}
\end{equation}
A single scan is therefore both one-sided and anisotropic. Writing the 2D coordinates $s = (i,j)$,
$t = (i',j')$ and $(\Delta_i, \Delta_j) = (i - i',\, j - j')$, the raster orderings of \eqref{eq:traversal_orders} give
\begin{equation}
d_1 = d_2 = \bigl| \Delta_i\, W + \Delta_j \bigr|,
\qquad
d_3 = d_4 = \bigl| \Delta_j\, H + \Delta_i \bigr|,
\end{equation}
where $d_n$, $n=1,2,3,4$ is the distance between $s$ and $t$ in different traverse orders.
So under $\pi_1$ alone a horizontal neighbor sits at ordering distance $1$ while a vertical neighbor sits at distance $d_1$ and is attenuated by $\rho^{d_1}$ rather than $\rho$, a result of the flattening, not of the underlying PDE field.

Summing the four directions in \eqref{eq:ssm} removes both one-sidedness and mitigates directional bias. Since $\pi_2 = N + 1 - \pi_1$ and $\pi_4 = N + 1 - \pi_3$, the four orderings form two reversed pairs, and for any ordered pair $(s,t)$ exactly one member of each pair satisfies the causality condition above. The combined scan orderings therefore permit a dependency path between every pair of locations. However, this structural connectivity does not guarantee a nonzero kernel value, since learned weights may vanish or contributions from different directions may cancel. The magnitude is governed by the shortest of the four orderings,
\begin{equation}
\bigl\|\kappa_{\mathrm{local}}(s,t)\bigr\|
\lesssim \rho^{\,d_\star(\Delta_i,\Delta_j)},
\qquad
d_\star
= \min\bigl(
    |\Delta_i W+\Delta_j|,
    |\Delta_j H+\Delta_i|
  \bigr).
\end{equation}
The distance $d_\star$ is invariant under simultaneous sign reversal, $(\Delta_i,\Delta_j)\mapsto(-\Delta_i,-\Delta_j)$. For square grids ($H=W$), it is also invariant under exchanging the coordinate offsets, $(\Delta_i,\Delta_j)\mapsto(\Delta_j,\Delta_i)$. In particular $d_\star = 1$ for all four axis-aligned nearest neighbors, whereas any single ordering places two of them at distance $W$ or $H$. The decay profile of the spatial branch is therefore balanced across both axes and both signs. We refer to this property as local isotropy. The property is exact on the coordinate axes but not for arbitrary displacements. For example, the offset $(1,1)$ has $d_\star=\min(W+1,H+1)$, whereas its Euclidean distance is $\sqrt{2}$ in grid units, so purely diagonal interactions remain attenuated in the spatial branch and are instead carried by the spectral branch and by the multi-resolution hierarchy.

\subsection{Spectral branch}
\paragraph{The spectral branch permits global interactions without recurrent distance decay.}
Starting from the integral-operator representation in \eqref{eq:kernel_integral}, we characterize the nonlinear spectral branch through its effective kernel:
\begin{equation}
\delta\mathcal{K}_{\mathrm{global}}(X)(s)
=
\int_{\Omega}
\kappa_{\mathrm{global}}(s,t)\,
\delta X(t)\,\mathrm{d}t,
\qquad
\kappa_{\mathrm{global}}(s,t)
=
\frac{\delta\mathcal{K}_{\mathrm{global}}(X)(s)}
     {\delta X(t)}.
\label{eq:spectral_local_kernel}
\end{equation}
As specified in \eqref{eq:lift}, the nonlinearities in \eqref{eq:blockmlp} and \eqref{eq:softshrink} act separately on the real and imaginary components. For a complex vector $z=x+iy$, consider the componentwise map $\phi_{\mathbb{C}}(z)=\phi(x)+i\phi(y)$. Define
\begin{equation}
    D_r=\operatorname{diag}\bigl(\phi'(x)\bigr),
    \qquad
    D_i=\operatorname{diag}\bigl(\phi'(y)\bigr),
\end{equation}
where $\phi'$ is evaluated elementwise. Its differential is
\begin{equation}
    \delta\phi_{\mathbb{C}}
    =D_r\,\delta x+iD_i\,\delta y
    =\frac{D_r+D_i}{2}\,\delta z
     +\frac{D_r-D_i}{2}\,\overline{\delta z}.
     \label{eq:complex_differential}
\end{equation}
The conjugate term is present whenever $D_r\neq D_i$, that is, when the derivatives evaluated at the real and imaginary components differ.

Suppressing the block index $\ell$ in \eqref{eq:blockmlp}, let $P_k$ and $Q_k$ denote the $\frac{D_r(k)+D_i(k)}{2}$ and $\frac{D_r(k)-D_i(k)}{2}$ of the real and imaginary ReLU derivative matrices, evaluated at $W_1Z(k)+b_1$ in \eqref{eq:complex_differential}. Similarly, let $R_k$ and $S_k$ denote the $\frac{D_r(k)+D_i(k)}{2}$ and $\frac{D_r(k)-D_i(k)}{2}$ of the soft-shrinkage derivative matrices, evaluated at $\tilde{Z}(k)$. The derivatives are $\operatorname{ReLU}'(u)=\operatorname{diag}[u>0]$ and $S_\lambda'(u)=\operatorname{diag}[|u|>\lambda]$ away from their thresholds.
Applying \eqref{eq:complex_differential} to \eqref{eq:blockmlp} gives
\begin{equation}
    \delta\tilde{Z}(k) = W_2P_kW_1\,\delta Z(k) + W_2Q_k\overline{W_1}\,
    \overline{\delta Z(k)}.
    \label{eq:blockmlp_differential}
\end{equation}
Differentiating \eqref{eq:softshrink} then yields
\begin{equation}
\begin{aligned}
    \delta\hat{Z}(k)
    &= R_k\,\delta\tilde{Z}(k)
     + S_k\,\overline{\delta\tilde{Z}(k)} \\
    &= A_k\,\delta Z(k)
     + B_k\,\overline{\delta Z(k)},
\end{aligned}
\label{eq:spectral_differential}
\end{equation}
where
\begin{equation}
    \begin{aligned}
        A_k &= R_kW_2P_kW_1 + S_k\overline{W_2}Q_kW_1,\\
        B_k &= R_kW_2Q_k\overline{W_1} + S_k\overline{W_2}P_k\overline{W_1}.
    \end{aligned}
    \label{eq:spectral_differential_coefficients}
\end{equation}
The coefficients depend on the input and frequency through the activation masks, although the learned weights are shared across frequencies. The biases $b_1,b_2$ affect these masks and are therefore retained in their evaluation. The formulas hold away from activation thresholds. At the thresholds, the derivatives are understood according to the automatic differentiation convention used by the implementation.

To recover the spatial kernel, we start from the \eqref{eq:kernel_integral} on the discrete grid $\hat{\Omega}=\mathbb{Z}_H\times\mathbb{Z}_W$:
\begin{equation}
\mathcal{K}(X)(s)=\sum_{t\in\hat{\Omega}}\kappa(s,t)\,X(t), \qquad s \in \hat{\Omega}.
\label{eq:discrete_afno_integral}
\end{equation}
We write $k \in \tilde{\Omega},\tilde{\Omega} = \mathbb{Z}_H\times\mathbb{Z}_W$ for a frequency index and $\langle k,s\rangle=\frac{k_1s_1}{H}+\frac{k_2s_2}{W}$, so that a field $X$ and its discrete
Fourier transform $Z=\mathcal{F}(X)$ are related by
\begin{equation}
Z(k)=\frac{1}{\sqrt{N}}\sum_{s\in\hat{\Omega}}X(s)\,e^{-2\pi i\langle k,s\rangle},
\qquad
X(s)=\frac{1}{\sqrt{N}}\sum_{k\in\tilde{\Omega}}Z(k)\,e^{2\pi i\langle k,s\rangle}.
\label{eq:dft}
\end{equation}

For the stored half-spectrum $\Omega_+=\mathbb{Z}_H\times\{0,\ldots,W_r-1\}$, let $\eta(k_2)=1$ on the columns $k_2=0$ and, when $W$ is even, $k_2=W/2$, and let $\eta(k_2)=2$ otherwise. These weights account for conjugate completion in the inverse
real FFT.

Substituting \eqref{eq:spectral_differential} and \eqref{eq:dft}
into the differential of \eqref{eq:ifft}, and exchanging
the spatial and frequency sums, gives
\begin{equation}
    \begin{aligned}
        \delta X_{\mathrm{global}}(s)
        =& \,\delta X_f(s)\\
        &+\sum_{t\in\hat{\Omega}}
        \left[
        \frac{1}{N}\operatorname{Re}
        \sum_{k\in\Omega_+}\eta(k_2)
        \left(
        A_ke^{2\pi i\langle k,s-t\rangle}
        +
        B_ke^{2\pi i\langle k,s+t\rangle}
        \right)
        \right]\delta X_f(t).
    \end{aligned}
    \label{eq:spectral_kernel_expansion}
\end{equation}
Consequently,
\begin{equation}
\begin{aligned}
\kappa_{\mathrm{global}}(s,t)
&= \delta_{s,t}I + a_f(s-t) + b_f(s+t) \\
&= \delta_{s,t}I + a_f(-\Delta) + b_f(2s+\Delta),
\end{aligned}
\label{eq:global_kernel}
\end{equation}
where
\begin{equation}
    \begin{aligned}
        a_{f}(u)
        &=
        \frac{1}{N}\operatorname{Re}
        \sum_{k\in\Omega_+}
        \eta(k_2)A_ke^{2\pi i\langle k,u\rangle}, \\
        b_{f}(u)
        &=
        \frac{1}{N}\operatorname{Re}
        \sum_{k\in\Omega_+}
        \eta(k_2)B_ke^{2\pi i\langle k,u\rangle}, \\
        \Delta &= t-s.
    \end{aligned}
    \label{eq:spectral_kernel_components}
\end{equation}
The identity term comes from the residual in \eqref{eq:ifft}. The direct term depends on $s-t$ and is a convolution, whereas the conjugate term depends on $s+t$ and is a reflected convolution: it convolves $b_f$ with the reflected perturbation $\delta X_f(-t)$. Thus, the effective kernel need not depend only on displacement.

Two properties follow. First, the Fourier sums in \eqref{eq:spectral_kernel_components} permit support over the entire domain, allowing distant locations to interact without propagation through intermediate spatial positions. Second, unlike the transition products of the spatial branch, these sums impose no explicit monotonic or exponential distance-decay constraint. The absence of such an architectural constraint does not guarantee constant or nonzero kernel magnitude at large distances. The learned interactions may be oscillatory and input-dependent, with the reflected term additionally introducing dependence on the query and source locations separately.

\subsection{SSF Block}
\paragraph{The SSF block combines both kernel families.} 
The two preceding analyses use different kernel representations to highlight the properties of each branch: the spatial analysis uses an input-dependent recurrence kernel for the SSM, whereas the spectral analysis uses a differential kernel to characterize perturbation propagation. For the forward-kernel interpretation below, we use $\hat{\kappa}_{\mathrm{local}}$ and $\hat{\kappa}_{\mathrm{global}}$ to denote input-dependent coefficient kernels of the complete spatial and spectral branches, respectively. In particular, $\hat{\kappa}_{\mathrm{local}}$ incorporates the operations surrounding the SSM recurrence, including convolution, normalization, gating, and projections, while $\hat{\kappa}_{\mathrm{global}}$ is distinct from its corresponding differential kernel $\kappa_{\mathrm{global}}$ derived above. The additive offsets are omitted from the expressions below. Here, $\mathcal{K}(X)$ denotes the sum of the branch outputs before the pointwise fusion layers.

Splitting the channels assigns the two parameterizations to complementary subspaces, and summing the two branch outputs superposes them on a shared set of channels. Writing the block input as $X = [X_s; X_f]$ with $X_s, X_f \in \mathbb{R}^{D}$ and $D = C/2$, the kernel acting on the full channel space is the block row
\begin{equation}
\kappa_{\mathrm{SSF}}(s,t)
=
\begin{bmatrix}
\hat{\kappa}_{\mathrm{local}}(s,t)
&
\hat{\kappa}_{\mathrm{global}}(s,t)
\end{bmatrix},
\label{eq:composite_kernel}
\end{equation}
\begin{equation}
\begin{split}
\mathcal{K}(X)(s)
&=
\int_{\Omega}
\begin{bmatrix}
\hat{\kappa}_{\mathrm{local}}(s,t)
&
\hat{\kappa}_{\mathrm{global}}(s,t)
\end{bmatrix}
\begin{bmatrix}
X_s(t) \\[2pt]
X_f(t)
\end{bmatrix}
\,\mathrm{d}t \\[4pt]
&=
\int_{\Omega}
\hat{\kappa}_{\mathrm{local}}(s,t)\,X_s(t)\,\mathrm{d}t
+
\int_{\Omega}
\hat{\kappa}_{\mathrm{global}}(s,t)\,X_f(t)\,\mathrm{d}t,
\end{split}
\label{eq:fusion_kernel_action}
\end{equation}
so each of the $D$ output channels draws on both kernel families. The summation fuses the two parameterizations. And the MLP in \eqref{eq:fusion} then acts pointwise, composing with \eqref{eq:composite_kernel} as a pointwise nonlinear map that lifts the result back to $C$ channels. 
The forward kernel $\hat{\kappa}_{\mathrm{local}}$ incorporates
the recurrence kernel $\kappa_{\mathrm{local}}$ and the surrounding spatial-branch operations, while $\kappa_{\mathrm{global}}$ describes the differential of the spectral map represented by $\hat{\kappa}_{\mathrm{global}}$. These connections motivate complementary biases toward scan-mediated local interactions and direct global mixing.
The spatial pathway supports input-adaptive interactions with attenuation along scan paths, whereas the spectral pathway permits direct domain-wide interactions without a recurrent distance-decay constraint. As shown in \eqref{eq:global_kernel}, the nonlinear spectral kernel may depend on both displacement and the query and source locations separately. It is not restricted to a translation-invariant convolution. These properties motivate explicit fusion of local and global interactions, without assuming that either branch exclusively represents one type of feature. The block combines the two pathways at a cost that remains near-linear in the sequence length.

\section{IMPLEMENTATION DETAILS}
\label{app:impl}
As shown in Figure~\ref{fig:overall}, the spatial-spectral fusion (SSF) is constructed as an encoder-decoder structure. The original input is patched, processed by multiple stages of SSF blocks, and downsampled to coarser resolution. Skip connections link each encoder level to its corresponding decoder level through ConvNeXt blocks \citep{convnet}, except at the bottleneck, where no skip connection is used. In the decoders, the data is assembled through upsampling and patch recovery. In the following section, we will introduce the implementation details of our model.

\subsection{Overlap Patch} 
\label{app:patch}
Patch partitioning creates the visual tokens. We extract overlapping
patches with a sliding window of kernel size $q = p + o$ and stride $p$,
where $o$ denotes the overlap in pixels. Given an input
$X_{\mathrm{in}} \in \mathbb{R}^{H \times W \times C}$, zero-padding is applied so
that the window count is $H^{(1)} \times W^{(1)}$ with $H^{(1)} = \lceil H/p \rceil$
and $W^{(1)} = \lceil W/p \rceil$. Let $(r_j, s_j)$ be the 2D index of the
$j$-th window, $j = 1, \dots, H^{(1)}W^{(1)}$. The corresponding window $y_j \in \mathbb{R}^{q \times q \times C}$ is
\begin{equation}
y_j(u,v,c')
=
X_{\mathrm{in}}\bigl(
  (r_j-1)p+u,\;
  (s_j-1)p+v,\;
  c'
\bigr),
\qquad
1 \le u,v \le q.
\end{equation}
Each patch is linearly embedded by a shared weight
$V \in \mathbb{R}^{C^{(1)} \times C \times q \times q}$ and bias $b \in \mathbb{R}^{C^{(1)}}$:
\begin{equation}
  v_j^{(c)} = \sum_{u=1}^{q} \sum_{v=1}^{q} \sum_{c'=1}^{C}
  V^{(c, c', u, v)} \, y_j(u, v, c') + b^{(c)},
\label{eq:patch-embed}
\end{equation}
where $c$ indexes the $C^{(1)}$ embedding channels. The embedding is then passed through a layer norm getting the result $h^{(1)} \in \mathbb{R}^{H^{(1)} \times W^{(1)} \times C^{(1)}}$.

\subsection{SSF-Net}
\paragraph{SSF Block}
\label{app:SSF-mlp}
At each resolution level in the U-shaped architecture, multiple SSF blocks are stacked. We index resolution levels by $\ell \in \{1,\dots,L_{\mathrm{res}}\}$, where $\ell = 1$ is the finest level and $\ell = L_{\mathrm{res}}$ the bottleneck. $L_{\mathrm{res}}$ denotes the number of levels. 

Each SSF block wraps the split--process--fuse map of \eqref{eq:split}-\ref{eq:fusion} in a residual connection. The details of the two branches are introduced in Appendix \ref{app:spatial_branch} and Appendix \ref{app:spectral_branch}. At resolution level $\ell$, the input $h^{(\ell)} \in \mathbb{R}^{H^{(\ell)} \times W^{(\ell)} \times C^{(\ell)}}$ is split into $h_s^{(\ell)} \in \mathbb{R}^{H^{(\ell)} \times W^{(\ell)} \times D^{(\ell)}}$ and $h_f^{(\ell)} \in \mathbb{R}^{H^{(\ell)} \times W^{(\ell)} \times D^{(\ell)}}$, $D^{(\ell)}=C^{(\ell)}/2$. $h_{\mathrm{sum}}^{(\ell)} \in \mathbb{R}^{H^{(\ell)} \times W^{(\ell)} \times D^{(\ell)}}$ is the layer-normalized sum of the spatial and spectral branch outputs, $f_{\mathrm{spatial}}(h_s^{(\ell)})$ and $f_{\mathrm{spectral}}(h_f^{(\ell)})$, respectively. The fusion MLP of \eqref{eq:fusion} is a two-layer pointwise network with GeLU activation  and four times the width of the latent dimension, applied identically at every spatial location:
\begin{equation}
\begin{gathered}
\operatorname{MLP}(h_{\mathrm{sum}}^{(\ell)})
=
\bigl(
W_2\bigl(\sigma(W_1 h_{\mathrm{sum}}^{(\ell)}+b_1)\bigr)
+b_2
\bigr), \\
W_1 \in \mathbb{R}^{D_h^{(\ell)} \times D^{(\ell)}},
\qquad
W_2 \in \mathbb{R}^{C^{(\ell)} \times D_h^{(\ell)}},
\end{gathered}
\end{equation}
where $\sigma$ is GeLU and $D_h^{(\ell)}=4 \cdot D^{(\ell)}$ the hidden width. It carries no spatial mixing, so all coupling between locations is produced by the two branches. Its role is to recombine the fused channels and to lift the branch width $D^{(\ell)}$ back to the block width $C^{(\ell)}$.

\paragraph{SSF Stage}
Let $\mathcal{M}^{(\ell,\ast)}_{i}$ denote $i$-th SSF block with residual at level $\ell$, where $\ast \in \{\mathrm{enc},\mathrm{dec}\}$ indicates whether the block belongs to the encoder or the decoder, and write $\operatorname{SSF}^{(\ell,\ast)}_{i}$ for the split--process--fuse map of
\eqref{eq:split}-\eqref{eq:fusion} instantiated at level $\ell$. Each block applies it residually,
\begin{equation}
\mathcal{M}_i^{(\ell,\ast)}(h_i^{(\ell)})
= h_i^{(\ell)} + \operatorname{SSF}_i^{(\ell,\ast)}(h_i^{(\ell)}).
\end{equation}
where $h_i^{(\ell)} \in \mathbb{R}^{H^{(\ell)} \times W^{(\ell)} \times C^{(\ell)}}$ denotes the input for $i$-th SSF block in $\ell$-th SSF Stage.
So a stage is a composition of residual blocks. The stage at level $\ell$ is
\begin{equation}
\mathcal{S}^{(\ell,\ast)}
= \mathcal{M}^{(\ell,\ast)}_{n^{(\ell)}} \circ
  \mathcal{M}^{(\ell,\ast)}_{n^{(\ell)}-1} \circ \cdots \circ
  \mathcal{M}^{(\ell,\ast)}_{1},
\qquad
\ell = 1,\dots,L_{\mathrm{res}},
\quad \ast \in \{\mathrm{enc},\mathrm{dec}\},
\end{equation}
where $n^{(\ell)}$ is the number of SSF blocks at level $\ell$. The encoder and decoder use the same depth $n_\ell$ at a given level, but $\mathcal{S}^{(\ell,\mathrm{enc})}$ and $\mathcal{S}^{(\ell,\mathrm{dec})}$ have independent parameters, as do stages at different levels.

\paragraph{Encoder, bottleneck and decoder.}
\label{app:multiresolution}
Write $\mathcal{D}^{(\ell)}$ for the downsampling operator from level $\ell$ to $\ell+1$, a convolution with kernel size $3$, stride $2$ and padding $1$ that
halves both spatial dimensions and doubles the channel width. $\mathcal{U}^{(\ell)}$
for the corresponding upsampling operator, a transposed convolution with kernel
size $2$ and stride $2$ that reverses both, and $\mathcal{G}^{(\ell)}$ for the skip
path at level $\ell$, a stack of $n_{\mathrm{skip}}^{(\ell)}$ ConvNeXt blocks in their
original form \citep{convnet} that leaves the width $C^{(\ell)}$ unchanged. With $h^{(1)}$ the output of the overlapping patch embedding (\eqref{eq:patch_embed}), the encoder proceeds as
\begin{equation}
e^{(\ell)} = \mathcal{S}^{(\ell,\mathrm{enc})}(h^{(\ell)}),
\qquad
h^{(\ell+1)} = \mathcal{D}^{(\ell)}(e^{(\ell)}),
\qquad
\ell = 1,\dots,L_{\mathrm{res}}-1 .
\end{equation}
The bottleneck is at the coarsest level and carries no skip connection,
\begin{equation}
d^{(L_{\mathrm{res}})}= \mathcal{S}^{(L_{\mathrm{res}},\mathrm{dec})}\mathcal{S}^{(L_{\mathrm{res}},\mathrm{enc})}(h^{(L_{\mathrm{res}})}),
\end{equation}
and the decoder ascends symmetrically, fusing the skip by addition before the stage,
\begin{equation}
d^{(\ell)}
= \mathcal{S}^{(\ell,\mathrm{dec})}\big(
\mathcal{U}^{(\ell)}(d^{(\ell+1)})
+ \mathcal{G}^{(\ell)}(e^{(\ell)})
\big),
\qquad
\ell = L_{\mathrm{res}}-1,\dots,1 .
\label{eq:decoder}
\end{equation}
The prediction is obtained by applying the de-patching operator of
Appendix~\ref{app:depatch} to $\mathbf{d}^{(1)}$. The total number of SSF blocks is $2\sum_{\ell=1}^{L_{\mathrm{res}}} n^{(\ell)}$.

\subsection{De-patching}
\label{app:depatch}
De-patching inverts the operation of Appendix~\ref{app:patch}, mapping the
finest decoder output back to a field on the original $H \times W$ grid with $C_{\mathrm{out}}$
physical channels. Re-indexing the patch index $j \in \{1,\dots,H^{(1)} \cdot W^{(1)}\}$ of
\eqref{eq:patch-embed} as a grid pair $(m,n)$, $m \in \{1,\dots,H^{(1)}\}, n  \in \{1,\dots,W^{(1)}\}$, let $d^{(c)}(m,n)$, $c = 1,\dots,C^{(1)}$, denote the decoder output at the finest resolution.

A transposed convolution with kernel size and stride both equal to $p$ scatters each
token onto a $p \times p$ block. Writing the pixel index as $i = (m-1)p + u$,
$j = (n-1)p + v$ with $u,v \in \{1,\dots,p\}$,
\begin{equation}
r^{(c)}(i,j) = \sum_{c'=1}^{C^{(1)}} V^{(c,c',u,v)}\, d(m,n,c') + b^{(c)},
\qquad
V \in \mathbb{R}^{C^{(1)} \times C^{(1)} \times p \times p}, \quad b \in \mathbb{R}^{C^{(1)}}.
\end{equation}
Unlike the embedding kernel of size $p+o$, this kernel has no overlap: each pixel is
written by exactly one token, so $r \in \mathbb{R}^{H \times W \times C^{(1)}}$ is
discontinuous at patch borders. A $3\times3$ convolution then projects back to the $n$
physical channels while mixing across those borders,
\begin{equation}
\begin{gathered}
u^{(c)}(i,j)
=
\sum_{\delta=-1}^{1}
\sum_{\varepsilon=-1}^{1}
\sum_{c'=1}^{C^{(1)}}
V'^{(c,c',\delta,\varepsilon)}
\,r(i+\delta,j+\varepsilon,c')
+\hat{b}^{(c)}, \\
V'
\in
\mathbb{R}^{C_{\mathrm{out}} \times C^{(1)} \times 3 \times 3},
\qquad
\hat{b} \in \mathbb{R}^{C_{\mathrm{out}}}.
\end{gathered}
\end{equation}
the out-of-range indices resolved by a padding rule
matching the boundary condition of the task. The asymmetry with the embedding is
deliberate: overlap on the encoder side aggregates neighbouring context into each
token, whereas an overlapping transposed convolution would give each output pixel a
spatially non-uniform number of contributions.

\subsection{SPATIAL BRANCH} 
\label{app:spatial_branch}
As shown in \eqref{eq:split}, each input $X \in \mathbb{R}^{H \times W \times C}$ is split into $X_s, X_f \in \mathbb{R}^{H \times W \times D}, D=C/2$ for the two branches. The multi-directional Mamba is used in spatial branch in the SSF blocks. 

\paragraph{Multi-directional traverse}
We define the traverse operator $\pi_n$ in \eqref{eq:traverse} in this paragraph. Let $\Omega_{\mathrm{pos}} = \{(i,j) : 1 \le i \le H,\ 1 \le j \le W\}$ denote the grid indices at the current resolution, with $N = HW$. Each traversal direction $n \in \{1,2,3,4\}$ is a bijection $\pi_n : \Omega_{\mathrm{pos}} \to \{1,\dots,N\}$ assigning to every grid point its position in the flattened sequence:
\begin{equation}
\begin{aligned}
\pi_1(i,j) &= (i-1)W + j, &\qquad \pi_2(i,j) &= N + 1 - \pi_1(i,j), \\
\pi_3(i,j) &= (j-1)H + i, &\qquad \pi_4(i,j) &= N + 1 - \pi_3(i,j),
\end{aligned}
\label{eq:traversal_orders}
\end{equation}
corresponding to row-major, reversed row-major, column-major, and reversed column-major orderings. Each ordering is a raster scan: every row (column) is read in the same direction. This keeps the ordering consistent with the reversal identities $\pi_2 = N+1-\pi_1$ and $\pi_4 = N+1-\pi_3$, so the four directions form two reversed pairs. The inverse maps $\pi_n^{-1}$ are obtained by the same formulas read backwards and are applied to restore the grid layout.

\paragraph{Selective SSM.} In a selective SSM, $\Delta$, $\mathbf{B}$ and $\mathbf{C}$ are input-dependent. Each scan direction $n \in \{1,2,3,4\}$ forms a group with its own state-space parameters $\mathbf{B}_n$ and $\mathbf{C}_n$, and each group is split into $M$ parallel heads that share $\mathbf{B}_n$ and $\mathbf{C}_n$ but carry their own timescale $\Delta_n^m$ and decay $\mathbf{A}_n^m$:
\begin{equation}
\begin{gathered}
\mathbf{A}_n^m = -\exp(\alpha_n^m), \\
\Delta_n,\mathbf{B}_n,\mathbf{C}_n
=\operatorname{Linear}_n(X_n), \\
\bar{\mathbf{A}}_n^m,\bar{\mathbf{B}}_n
=\operatorname{discretize}\bigl(
  \Delta_n^m,\mathbf{A}_n^m,\mathbf{B}_n\bigr),
\qquad m=1,\dots,M, \\
X_{\mathrm{ssm}}=
\sum_{n=1}^{4}
\pi_n^{-1}\Bigl(
  \operatorname{Concat}\bigl(
    \operatorname{SSM}(
      \bar{\mathbf{A}}_n^1,
      \bar{\mathbf{B}}_n,
      \mathbf{C}_n
    )(X_n^1),
    \ldots,
    \operatorname{SSM}(
      \bar{\mathbf{A}}_n^M,
      \bar{\mathbf{B}}_n,
      \mathbf{C}_n
    )(X_n^M)
  \bigr)
\Bigr).
\end{gathered}
\end{equation}
where $\alpha^m_n$ is a learnable parameter, $\operatorname{SSM}(\bar{\mathbf{A}}_n^m, \bar{\mathbf{B}}_n, \mathbf{C}_n)(X_n^m)$  denotes the recurrence \eqref{eq:conv_ssm_dependent} run with $(\bar{\mathbf{A}}_n^m, \bar{\mathbf{B}}_n, \mathbf{C}_n)$ on the $m$-th channel slice of $X_n$, and $\pi_n^{-1}$ is the inverse of the traversal $\pi_n$. Only $\Delta_n$, $\mathbf{B}_n$ and $\mathbf{C}_n$ are produced from the input. $\mathbf{A}_n^m$ is a learned scalar. Within a direction, all heads share the same $(\mathbf{B}_n, \mathbf{C}_n)$ and differ only in $\mathbf{A}_n^m$.

\subsection{Spectral Branch: Nonlinearity on the complex domain}
\label{app:spectral_branch}
Both the ReLU of \eqref{eq:blockmlp} and the soft-shrinkage of \eqref{eq:softshrink} act in the spectral domain, where the features are complex-valued. Since complex field $\mathbb{C}$ admits no natural ordering, neither is defined on a complex value as a whole. We therefore identify $\mathbb{C}$ with $\mathbb{R}^2$ and apply both nonlinearities to the real and imaginary parts independently. Writing
\begin{equation}
\mathcal{R}: \mathbb{C} \to \mathbb{R}^2, \qquad
\mathcal{R}(a + i\,b) = (a,\, b),
\end{equation}
for this identification, a real-valued nonlinearity $\phi$ is lifted to the complex
domain as
\begin{equation}
\phi_{\mathbb{C}}(z)
=
\mathcal{R}^{-1}\!\bigl(\phi(\mathcal{R}(z))\bigr),
\qquad \text{i.e.} \qquad
\phi_{\mathbb{C}}(a+i\,b)
=
\phi(a)+i\,\phi(b),
\label{eq:lift}
\end{equation}
where $\phi$ acts elementwise. We use \eqref{eq:lift} for both
$\phi = \operatorname{ReLU}$ and $\phi = S_\lambda$.

\paragraph{Complex linear layer.}
Writing a complex weight as $W = W_r + i\,W_i$ with $W_r, W_i$ real, the product
$Wz$ for $z = x_r + i\,x_i$ expands to
\begin{equation}
\mathcal{R}(Wz) \;=\;
\begin{pmatrix} W_r & -W_i \\ W_i & \phantom{-}W_r \end{pmatrix}
\begin{pmatrix} x_r \\ x_i \end{pmatrix},
\label{eq:cmul}
\end{equation}
so a complex linear layer is realised as a real linear layer with the structure
of \eqref{eq:cmul}, and both of its outputs are real-valued. Composing
\eqref{eq:cmul} with \eqref{eq:lift}, one layer of the block-diagonal complex MLP in
\eqref{eq:blockmlp} is computed as
\begin{equation}
\mathcal{R}\bigl(\operatorname{ReLU}_{\mathbb{C}}(Wz)\bigr)
=
\operatorname{ReLU}\!\left(
\begin{pmatrix}
W_r & -W_i \\
W_i & W_r
\end{pmatrix}
\begin{pmatrix}
x_r \\
x_i
\end{pmatrix}
\right).
\end{equation}
\paragraph{Soft-shrinkage.}
The same treatment applies to the sparsification of \eqref{eq:softshrink}. Writing
$\tilde{z} = \tilde{z}_r + i\,\tilde{z}_i$ for an entry of $\tilde{Z}$, the threshold is applied to each part separately,
\begin{equation}
\hat{z} \;=\; S_\lambda(\tilde{z}_r) \;+\; i\, S_\lambda(\tilde{z}_i),
\qquad
S_\lambda(x) = \operatorname{sign}(x)\max\{|x| - \lambda,\, 0\},
\label{eq:cshrink}
\end{equation}
with $\lambda$ a fixed threshold, shared across all levels and channels and not learned. An entry is zeroed if and only if the absolute values of both its real and imaginary parts are at most $\lambda$. For nonzero outputs, the operation reduces the magnitude
and may also change the phase.

\section{Experiments and hyperparameters}
\label{app:experiments}

\subsection{Dataset Preprocessing and Input Construction}
\label{app:data}

\paragraph{PDE GYM datasets}
As shown in Table~\ref{tab:pdegym_datasets}, we use the public PDE GYM datasets introduced by \citet{poseidon}, including Poisson-Gauss, NS-PwC, NS-SL, NS-Tracer-PwC, FNS-KF, CE-RP, Wave-Gauss, and ACE. We use the original train/validation/test splits, spatial resolutions, temporal sampling intervals, and physical variables. Unless otherwise stated, preprocessing and normalization follow the official PDE GYM protocol. For time-dependent tasks, consecutive states $(u_t,u_{t+1})$ are used to construct one-step training pairs. For tasks with static conditioning fields, the normalized static field is concatenated with the dynamic state along the channel dimension and supplied to the model at every rollout step. In particular, Wave-Gauss uses the spatially varying wave-speed field, whereas FNS-KF uses the prescribed forcing field. The model predicts only the dynamic solution channels, while the static conditioning channels are preserved throughout autoregressive rollout.

\begin{table*}[t]
    \centering
    \caption{
    PDE GYM and APEBench-generated datasets used in our experiments. TI and TD denote time-independent and time-dependent tasks, respectively. All datasets are represented on $128\times128$ Cartesian grids.
    }
    \label{tab:pdegym_datasets}
    \resizebox{\textwidth}{!}{%
    \begin{tabular}{lllccc}
        \toprule
        Dataset & Governing equation & Type & Model input
        & Prediction target & Static conditioning \\
        \midrule
        \multicolumn{6}{l}{\textit{PDE GYM}} \\
        \midrule
        Poisson-Gauss & Poisson & TI
        & $f$ & $u$ & -- \\

        NS-PwC & Incompressible Navier--Stokes & TD
        & $(u_x^t,u_y^t)$
        & $(u_x^{t+1},u_y^{t+1})$ & -- \\

        NS-Tracer-PwC & Navier--Stokes with passive tracer & TD
        & $(u_x^t,u_y^t,c_{\mathrm{tr}}^t)$
        & $(u_x^{t+1},u_y^{t+1},c_{\mathrm{tr}}^{t+1})$ & -- \\

        NS-SL & Incompressible Navier--Stokes & TD
        & $(u_x^t,u_y^t)$
        & $(u_x^{t+1},u_y^{t+1})$ & -- \\

        FNS-KF & Forced incompressible Navier--Stokes & TD
        & $(u_x^t,u_y^t,g)$
        & $(u_x^{t+1},u_y^{t+1})$ & $g$ \\

        CE-RP & Compressible Euler & TD
        & $(\rho^t,u_x^t,u_y^t,p^t)$
        & $(\rho^{t+1},u_x^{t+1},u_y^{t+1},p^{t+1})$ & -- \\

        Wave-Gauss & Acoustic wave equation & TD
        & $(u^t,c_{\mathrm{w}})$
        & $u^{t+1}$ & $c_{\mathrm{w}}$ \\

        ACE & Allen--Cahn & TD
        & $u^t$ & $u^{t+1}$ & -- \\

        \midrule
        \multicolumn{6}{l}{\textit{APEBench-generated}} \\
        \midrule
        Kolm-Smooth & Forced Navier--Stokes (vorticity form) & TD
        & $u^t$ & $u^{t+1}$ & -- \\

        Kolm-Turb & Forced Navier--Stokes (vorticity form) & TD
        & $u^t$ & $u^{t+1}$ & -- \\

        KS-Strong & Kuramoto--Sivashinsky & TD
        & $u^t$ & $u^{t+1}$ & -- \\

        KS-Weak & Kuramoto--Sivashinsky & TD
        & $u^t$ & $u^{t+1}$ & -- \\
        \bottomrule
    \end{tabular}%
    }

    \vspace{2pt}
    \parbox{\textwidth}{\footnotesize
    \textit{Notation.}
    $f$ and $u$ denote the source and solution fields in the Poisson equation, respectively. $(u_x,u_y)$ are the two velocity components. $\rho$ and $p$ denote fluid density and pressure. $g$ is the prescribed forcing field supplied as a static input channel in FNS-KF. $c_{\mathrm{tr}}$ is the passive-tracer concentration, and $c_{\mathrm{w}}$ is the spatially varying wave speed. For Kolm-Smooth and Kolm-Turb, $u$ denotes scalar vorticity. For KS-Strong and KS-Weak, it denotes the scalar state field. The prescribed forcing in the Kolmogorov-flow datasets is included in the data-generating dynamics but is not supplied as an additional model input. Superscripts $t$ and $t+1$ indicate consecutive time steps.
    }
\end{table*}

\paragraph{APEBench-generated datasets}
We additionally generate four two-dimensional long-horizon datasets using APEBench \citep{koehler2024apebench}: Kolm-Turb, Kolm-Smooth, KS-Strong, and KS-Weak. Each dataset contains 4000 trajectories on a $128\times128$ grid, divided into 3,700 training, 200 validation, and 100 test trajectories. The first 3,900 trajectories are generated using the APEBench training-data generator, while the remaining 100
test trajectories are generated separately using its test-data generator. Each trajectory contains 101 states, corresponding to 100 state transitions, following 500 numerical warm-up steps.

\subparagraph{Kolmogorov-flow datasets.}
We generate the Kolm-Turb and Kolm-Smooth datasets by solving the two-dimensional incompressible Navier--Stokes equation in vorticity form on a periodic domain:
\begin{equation}
    \frac{\partial u}{\partial t}
    =
    -b
    \left(
        \begin{bmatrix}1\\-1\end{bmatrix}
        \odot \nabla(\Delta^{-1}u)
    \right)
    \cdot\nabla u
    +\nu\Delta u
    +\lambda u
    - k \cos\left(k\frac{2\pi}{L}y\right),
    \label{eq:kolmogorov_flow}
\end{equation}
where $u=u(\mathbf{x},t)$ denotes the scalar vorticity field. The coefficient $b$ controls the strength of nonlinear vorticity advection, $\nu$ is the kinematic viscosity, and $\lambda<0$ is a linear drag coefficient that removes energy accumulated in the low-frequency modes. The last term is the Kolmogorov forcing, where $k$ is the injection wavenumber and $L$ is the side length of the periodic spatial domain. 

Both Kolm-Turb and Kolm-Smooth use the same forcing wavenumber $k=4$, forcing scale $1.0$, stored time interval $\Delta t=0.1$, and 20 numerical substeps per stored transition. The two configurations differ only in the viscosity $\nu$ and linear
drag coefficient $\lambda$. Kolm-Turb uses $\nu=1.0\times10^{-2}$ and $\lambda=-0.1$ to produce a more turbulent and challenging long-horizon forecasting regime, whereas Kolm-Smooth uses $\nu=1.5\times10^{-2}$ and $\lambda=-0.15$ exhibiting smoother spatial fields and less intense spatiotemporal variation.

\subparagraph{Kuramoto--Sivashinsky datasets.}
KS-Strong and KS-Weak are generated from the Kuramoto-Sivashinsky equation:
\begin{equation}
    \frac{\partial u}{\partial t}
    =
    -\frac{b}{2}\lVert\nabla u\rVert_2^2
    -\nu\nabla\cdot\nabla u
    -\zeta\,\mathbf{1}^{\top}\nabla^4u,
    \label{eq:ks_2d}
\end{equation}
where $u=u(\mathbf{x},t)$ is the scalar state field. The coefficient $b$ controls the nonlinear gradient-norm term. $\nu$ controls the destabilizing second-order term, and $\zeta$ controls the stabilizing fourth-order hyperdiffusion term. The competition between the second-order instability, fourth-order dissipation, and nonlinear gradient interaction gives rise to complex spatiotemporal dynamics.

Although Equation~\eqref{eq:ks_2d} is expressed using the physical coefficients $b$, $\nu$, and $\zeta$, our datasets are generated through the difficulty-based interface of APEBench. Both KS-Strong and KS-Weak use the same gradient-norm difficulty
coefficient $\delta_2=-6.0$ and fourth-order hyperdiffusion difficulty coefficient $\gamma_4=-15.0$. The two configurations differ only in the second-order diffusion difficulty coefficient $\gamma_2$. KS-Strong uses $\gamma_2=-1.2$, resulting in a stronger destabilizing anti-diffusion effect and producing more intense spatiotemporal
variations, whereas KS-Weak uses $\gamma_2=-0.8$, which weakens the second-order instability and produces comparatively milder and smoother dynamics.

\subsection{Training and metrics}
\label{app:evaluation_metrics}
We train a one-step operator $\mathcal{S}_\theta: u_t \mapsto u_{t+1}$ with the standard relative $L^1$ loss. For Drift-Net, we use frequency-weighted loss proposed in \citet{drift-net}. At test stage, $\mathcal{S}_\theta$ runs autoregressively to generate the full trajectory $\{u_t\}_{t=1}^{T}$. We report step-averaged relative $L_1$ error $\mathcal{E}_{\mathrm{step}}$, full-rollout relative $L_1$ error $\mathcal{E}_{\mathrm{full}}$ and final-step relative $L_1$ error $\mathcal{E}_{\mathrm{final}}$:
\begin{equation}
\mathcal{E}_{\mathrm{step}}
=
\frac{1}{N_{\mathrm{test}}T}
\sum_{i=1}^{N_{\mathrm{test}}}
\sum_{t=1}^{T}
\frac{\left\|\widehat{u}_i^t-u_i^t\right\|_1}
     {\left\|u_i^t\right\|_1},
\label{eq:test_step_l1}
\end{equation}
\begin{equation}
\mathcal{E}_{\mathrm{full}}
=
\frac{1}{N_{\mathrm{test}}}
\sum_{i=1}^{N_{\mathrm{test}}}
\frac{\sum_{t=1}^{T}
      \left\|\widehat{u}_i^t-u_i^t\right\|_1}
     {\sum_{t=1}^{T}
      \left\|u_i^t\right\|_1},
\label{eq:test_full_l1}
\end{equation}
\begin{equation}
\mathcal{E}_{\mathrm{final}}
=
\frac{1}{N_{\mathrm{test}}}
\sum_{i=1}^{N_{\mathrm{test}}}
\frac{\left\|\widehat{u}_i^T-u_i^T\right\|_1}
     {\left\|u_i^T\right\|_1},
\label{eq:test_final_l1}
\end{equation}

where $u_i^t$ and $\widehat{u}_i^t$ denote the ground-truth field and the corresponding autoregressive prediction at forecast step $t$ of the $i$-th test trajectory. $N_{\mathrm{test}}$ is the number of test trajectories, and $T$ is the number of forecast steps. Each rollout starts from the ground-truth initial state $\widehat{u}_i^0=u_i^0$.

The step-averaged error assigns equal importance to every forecast step. The full-rollout error evaluates the predicted trajectory as a whole and accounts for the relative contribution of states with different magnitudes. The final-step error focuses on the longest prediction horizon and is particularly sensitive to accumulated autoregressive errors and instability. Reporting all three metrics therefore provides a more complete assessment of average per-step accuracy, overall trajectory fidelity, and long-term rollout stability than any individual metric alone.

For tasks with multiple quantities of interest (QoIs), we follow the setting of \citet{drift-net} computing the relative $L_1$ error for each QoI and take an unweighted average.

\subsection{Training and Optimization}
\label{app:training}
\paragraph{Training objective and batching.}
For all time-dependent datasets, we train a one-step neural operator $\mathcal{S}_{\theta}:u_t\mapsto u_{t+1}$ using single-step teacher forcing. Each pair of consecutive ground-truth states $(u_t,u_{t+1})$ constitutes one supervised training example. 

\revised{Unlike the Drift-Net \citep{drift-net}, which samples individual trajectory--time pairs and allows different temporal positions to be mixed within a batch, we organize training batches by complete trajectories and process their adjacent transitions sequentially, performing a separate optimizer update for each time index. Thus, each update uses transitions from different trajectories at the same temporal position. All adjacent transitions in the sampled trajectory batch are processed before moving to the next batch. }

\revised{This organization systematically uses all adjacent transitions from each sampled trajectory batch, with a fixed prediction interval matching the one-step forecasting task. Grouping transitions by temporal position also avoids mixing different stages of trajectory evolution within an optimizer update and provides a consistent batching protocol across the compared architectures.}

Unless otherwise stated, models are optimized using the relative $L_1$ loss described in Section~\ref{app:experiments}. For multi-component datasets, the loss is computed separately for each predefined group of physical variables and then averaged across groups, preventing variables with larger channel counts from dominating the objective. Drift-Net instead uses its original frequency-weighted loss, following the configuration proposed in \citet{drift-net}. For the time-independent Poisson-Gauss task, the model is trained directly to predict the steady-state solution $u$ from the source field $f$.

\paragraph{Optimization.}
We optimize all models using AdamW. Unless an architecture-specific configuration is stated otherwise, we use an initial learning rate of $5\times10^{-4}$, a weight decay of $10^{-6}$, and a batch size of 40. Gradients are clipped to a maximum $\ell_2$ norm of 5. For SSF-Net, FNO, PDE-Transformer, and scOT, the learning rate is linearly
increased from $10^{-3}$ of its initial value to the full initial learning rate during the first two epochs, followed by cosine annealing to a minimum learning rate of $10^{-6}$. CNO uses a batch size of 32 and a step-decay schedule that multiplies the learning rate by $0.9$ every 10 epochs. Drift-Net uses its original initial learning rate of $3\times10^{-4}$ together with the same linear-warmup and cosine-decay schedule.

\paragraph{Epoch budget and checkpoint selection.}
All methods are assigned the same maximum number of training epochs on each dataset. We train for 20 epochs on NS-SL, 100 epochs on Poisson-Gauss, and 40 epochs on all remaining benchmarks. At the end of each epoch, we evaluate the model on the validation split. For time-dependent tasks, checkpoint selection is based on the
full-rollout relative $L_1$ validation error. For Poisson-Gauss, it is based on the relative $L_1$ error of the predicted steady-state solution. The test set is not used for optimization or checkpoint selection.

\paragraph{Hardware and precision.}
All experiments are implemented in PyTorch and conducted on a single NVIDIA H200 GPU in full single precision (FP32), without mixed-precision or distributed training.

\subsection{Model Configurations}
\label{app:model_configurations}
\revised{The model configurations of FNO, CNO, scOT follows the setting in Poseidon \citep{poseidon}, and we use the same setting of Drift-Net and PDE-Transformer as their original paper \citep{drift-net, pde-transformer}.} All models are trained from scratch separately on each dataset. For a given task, the input and output dimensions are adjusted according to the corresponding physical variables listed in Table~\ref{tab:pdegym_datasets}, while the remaining architectural configuration is kept fixed across datasets.

\paragraph{SSF-Net.}
The input field is embedded using a convolutional projection with patch size $2$ and base feature dimension $48$. The model contains $4$ resolution levels with $2$ SSF blocks at each level, resulting in the depth configuration $(2,2,2,2)$. The channel dimension is doubled after each encoder downsampling stage and symmetrically reduced in the decoder. Both downsampling and upsampling are implemented using learnable convolutional operators with kernel size $3$, stride $2$ and padding $1$. Skip connections link the corresponding encoder and decoder levels, with $(2,2,2,0)$ ConvNeXt blocks applied to the skip pathways from the shallowest to the deepest level, respectively. The resulting model contains approximately $11\text{M}$ trainable parameters.

\paragraph{FNO.}
Our FNO baseline follows the standard two-dimensional Fourier Neural Operator architecture \citep{fno}. The input is lifted to a hidden width of $96$ and processed by $5$ Fourier layers. Each Fourier layer retains $20$ modes along each spatial dimension and combines the spectral convolution with a pointwise linear transformation. The final features are projected to the task-specific output channels. This configuration contains approximately $36\text{M}$ trainable parameters.

\paragraph{CNO.}
We use the two-dimensional Convolutional Neural Operator \citep{cno} with $4$ encoder and decoder resolution levels. The channel multiplier is set to $32$, with $8$ residual blocks at each resolution level and $8$ additional residual blocks in the bottleneck. The model uses $3\times3$ convolutional kernels and CNO-specific filtered activation functions. This configuration contains approximately $18\text{M}$ trainable parameters.

\paragraph{PDE-Transformer.}
Considering we train the model on a single dataset from scratch, we only use the PDE-Transformer (MC)\citep{pde-transformer} as the baseline. The PDE-Transformer baseline uses a 5-level multiscale encoder--decoder with patch size $4$, base embedding dimension $96$ and block depth $(2,5,8,5,2)$. The resulting model contains approximately $33\text{M}$ trainable parameters.

\paragraph{scOT.}
We use the scOT architecture introduced in Poseidon \citep{poseidon}. The model employs a 4-level multiscale encoder--decoder with Swin-style window attention, patch size $4$, base embedding dimension $48$ and $4$ blocks at each resolution level, giving the depth configuration $(4,4,4,4)$. Skip connections link the corresponding encoder and decoder levels, with $(2,2,2,0)$ ConvNeXt blocks. This configuration contains approximately $20\text{M}$ trainable parameters.

\paragraph{Drift-Net.}
Our Drift-Net baseline follows the spatial--spectral architecture of \citet{drift-net}. It uses a 4-level encoder--decoder with patch size $4$, base embedding dimension $48$ and block depths $(4,4,4,4)$. Each block combines an image-space branch  (ConvNeXt-style block with depthwise $3 \times 3$ convolution, pointwise $1 \times 1$ convolution) with a spectral branch for frequency-domain feature mixing. We retain the original frequency-weighted training objective of Drift-Net. The resulting configuration contains approximately $17\text{M}$ trainable parameters.

\section{Additional ablation study}
\subsection{Branch Ablations at different Model Sizes}
\label{app:capacity_ablation}
The ablations in Table~\ref{tab:ablation} retain the same backbone configuration, including the number of resolution levels, block depths, stage-wise channel dimensions, patch size, and skip connections. In each single-branch variant, the retained branch is widened to preserve the block width. Because the spatial and spectral branches have different parameterizations, this yields different parameter counts. These experiments therefore compare branch choices within a fixed backbone rather than under a matched parameter budget.

To complement this comparison, we adjust the base embedding dimension of the single-branch variants, increasing the size of the spectral-only model and reducing that of the spatial-only model. The remaining architectural and training settings are unchanged. The resulting models contain approximately 10M and 8M parameters, respectively, compared with 11M for the full model. These configurations provide a comparison at similar, although not identical, parameter counts.

\begin{table}[t]
    \centering
    \caption{
        \revised{
        Branch ablations on ACE and FNS-KF with fixed-backbone
        and adjusted-width configurations.
        Each entry reports
        $\mathcal{E}_{\mathrm{step}}$,
        $\mathcal{E}_{\mathrm{full}}$, and
        $\mathcal{E}_{\mathrm{final}}$
        in percent; lower is better.
        The best and second-best results for each metric within
        each dataset are shown in bold and underlined,
        respectively, including ties.
        Parameter counts are approximate.
        }
    }
    \label{tab:capacity_ablation}
    \small
    \begin{tabular}{lccc}
        \hline
        Variant & Params & ACE & FNS-KF \\
        \hline
        Spectral only (fixed backbone)
        & 9M
        & 0.021/0.023/0.033
        & 0.115/0.115/0.188 \\

        Spatial only (fixed backbone)
        & 16M
        & \textbf{0.012}/\textbf{0.014}/\textbf{0.026}
        & \textbf{0.098}/\textbf{0.098}/\underline{0.168} \\
        \hline
        Spectral only (adjusted width)
        & 10M
        & 0.022/0.024/0.034
        & 0.122/0.122/0.206 \\

        Spatial only (adjusted width)
        & 8M
        & 0.022/0.023/\underline{0.032}
        & \underline{0.109}/\underline{0.109}/0.172 \\

        Full SSF-Net
        & 11M
        & \underline{0.020}/\underline{0.021}/0.033
        & \textbf{0.098}/\textbf{0.098}/\textbf{0.157} \\
        \hline
    \end{tabular}
\end{table}

As shown in Table~\ref{tab:capacity_ablation}, the full model outperforms both adjusted-width single-branch variants on all three metrics on FNS-KF. On ACE, it achieves lower step-averaged and full-rollout errors, while the 8M spatial-only model achieves a slightly lower final-step error (0.032\% versus 0.033\%). The 16M spatial-only model remains the most accurate configuration on ACE among those evaluated. The advantage of the wider spatial-only model on ACE does not persist across all metrics after its capacity is reduced. Moreover, increasing the spectral-only model from approximately 9M to 10M parameters does not improve its errors in these experiments. Overall, the results support the full model as a competitive configuration at the evaluated model sizes, while indicating that branch composition and capacity must be considered jointly.

\subsection{Comparison with a Convolutional Spatial Branch}
\label{app:conv_spatial_ablation}
\begin{table}[t]
    \centering
    \caption{
    \revised{
        Comparison of Mamba-based and convolutional spatial
        branches on ACE and FNS-KF.
        Each entry reports
        $\mathcal{E}_{\mathrm{step}}/
         \mathcal{E}_{\mathrm{full}}/
         \mathcal{E}_{\mathrm{final}}$ in percent.
        Lower is better.
        Both models contain approximately 11M parameters.
        The lower error for each metric is shown in bold.
        }
    }
    \label{tab:conv_spatial_ablation}
    \small
    \setlength{\tabcolsep}{5pt}
    \begin{tabular}{lcc}
        \toprule
        Spatial branch & ACE & FNS-KF \\
        \midrule
        ConvNeXt
        & 0.022/0.024/0.035
        & 0.108/0.108/0.175 \\

        Multi-directional Mamba (SSF-Net)
        & \textbf{0.020}/\textbf{0.021}/\textbf{0.033}
        & \textbf{0.098}/\textbf{0.098}/\textbf{0.157} \\
        \bottomrule
    \end{tabular}
\end{table}

To further explore the contribution of the Mamba-based spatial pathway, we replace it with a ConvNeXt-style convolutional alternative while retaining the spectral pathway, fusion operation, patch embedding, multiresolution backbone, and training protocol. Both configurations contain approximately 11M parameters. This provides a comparison at similar parameter counts, although it does not establish an identical computational budget.

As shown in Table~\ref{tab:conv_spatial_ablation}, the Mamba-based spatial branch achieves lower errors under all three metrics on both tasks. The final-step error decreases from 0.035\% to 0.033\% on ACE and from 0.175\% to 0.157\% on FNS-KF. These results support the contribution of the Mamba-based spatial pathway within the evaluated SSF architecture, with a larger improvement on FNS-KF.

\section{Inference Efficiency}
\label{app:inference_efficiency}
\begin{table}[t]
    \centering
    \caption{
    \revised{
        Inference efficiency on FNS-KF with a batch size of 40.
        Latency is reported in $10^{-4}$ seconds per sample,
        and throughput in samples per second.
        Lower latency and higher throughput are better.
        }
    }
    \label{tab:inference_efficiency}
    \small
    \setlength{\tabcolsep}{6pt}
    \begin{tabular}{lcc}
        \toprule
        Model & Latency & Throughput \\
        \midrule
        FNO                  & 2.93  & 164.25 \\
        CNO                  & 9.29  & 53.03  \\
        PDE-Transformer      & 6.30  & 77.94  \\
        scOT                 & 4.64  & 105.02 \\
        Drift-Net            & 4.89  & 99.86  \\
        SSF-Net (patch size 2) & 10.05 & 49.13 \\
        SSF-Net (patch size 4) & 6.73  & 72.97 \\
        \bottomrule
    \end{tabular}
\end{table}

Table~\ref{tab:inference_efficiency} compares the inference latency and throughput of SSF-Net and the evaluated baselines on FNS-KF with a batch size of 40. We include both patch-size configurations of SSF-Net to examine the computational effect of tokenization.

SSF-Net with patch size 2 has higher inference latency and lower throughput than the compared baselines, despite having fewer parameters. Increasing its patch size to 4 reduces the initial token count by a factor of four and improves inference throughput from 49.13 to 72.97 samples per second, at the cost of higher prediction errors (Table~\ref{tab:ablation}). These measurements illustrate the trade-off between prediction accuracy and practical inference efficiency. They also show that the near-linear asymptotic complexity of SSF-Net does not guarantee lower wall-clock inference cost.

\section{Model Analysis}
\label{app:model_analysis}
\begin{figure}[t]
    \centering
    \includegraphics[width=\linewidth]
    {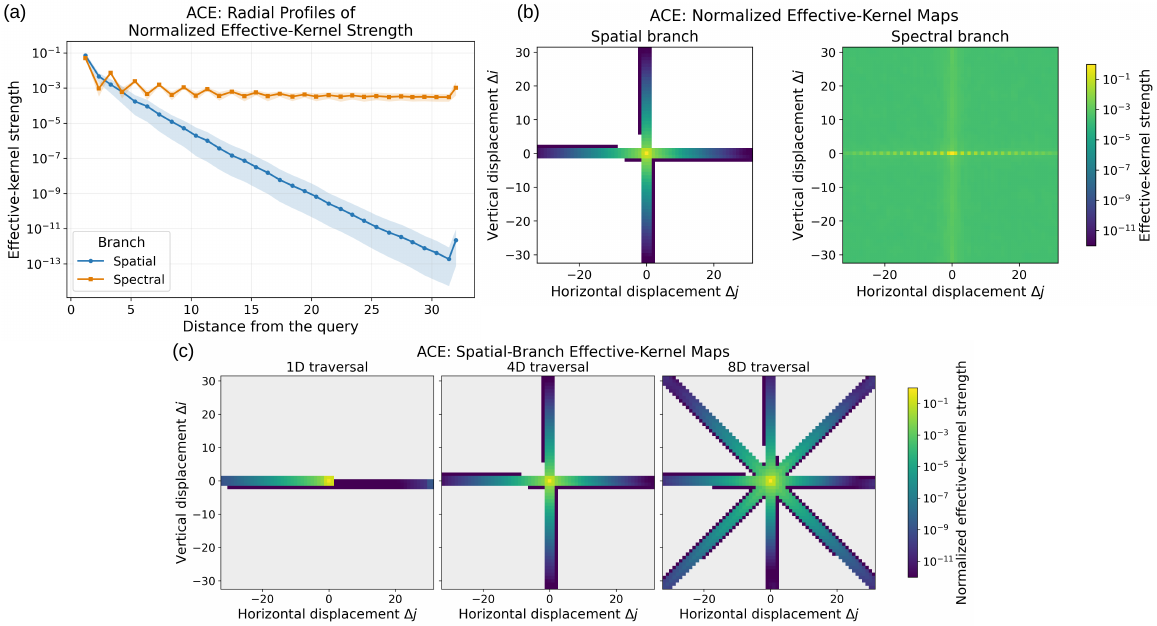}
    \caption{
    Effective-kernel analysis on ACE at the first encoder block. 
    (a) radially averaged normalized kernel strength, showing sample means and 2.5th--97.5th percentile bands. 
    (b) sample-averaged normalized kernel strength of the spatial and spectral branches of SSF-Net. 
    (c) spatial-branch maps with one, four, and eight traversal directions. All panels share a logarithmic color scale, with gray indicating zero.
    }
    \label{fig:ace-effective-kernel}
\end{figure}

\subsection{Effective-Kernel definition}
\paragraph{Measurement setup.}
For the effective-kernele strength in Figure~\ref{fig:effective-kernel}, Figure~\ref{fig:FNS-KF-effective-kernel} and Figure~\ref{fig:ace-effective-kernel}, we use 16 samples at the initial stored time index and process each sample separately. A forward hook captures the input features of the first block in the first encoder stage. We split the captured features into the spatial and spectral channel groups and evaluate each branch independently.  For other baselines in Figure~\ref{fig:effective-kernel}, we only capture the feature in the first encoder stage without split. The query is the central feature location $q$.

\paragraph{Jacobian-based kernel strength.}

Let $G_b$ denote the implemented branch map and $X_b^{(n)}$ its captured input for sample $n$, where $b\in\{\mathrm{spatial},\mathrm{spectral}\}$. For a source location $x$, we compute
\begin{equation}
    J_b^{(n)}(q,x)
    =\left.\frac{\partial G_b(X_b)(q)}{\partial X_b(x)}\right|_{X_b=X_b^{(n)}},
    \qquad
    K_b^{(n)}(q,x)=\left\|J_b^{(n)}(q,x)\right\|_F.
    \label{eq:diagnostic_kernel}
\end{equation}
The Frobenius norm sums squared derivatives over all input and output channels. We compute it by reverse-mode automatic differentiation, enumerating every output-channel basis vector at $q$ and summing the squared input gradients. Each sample and branch is normalized separately using the entire feature grid $\Omega_h$:
\begin{equation}
    \widetilde K_b^{(n)}(q,x)
    =\frac{K_b^{(n)}(q,x)}
    {\left(\sum_{x'\in\Omega_h}K_b^{(n)}(q,x')^2\right)^{1/2}},
    \qquad
    M_b(\delta)=\frac{1}{N}\sum_{n=1}^{N}
    \widetilde K_b^{(n)}(q,q+\delta),
    \label{eq:diagnostic_normalization}
\end{equation}
where $N=16$ and $\delta$ denotes displacement from the query.

\paragraph{Radial profiles.}
\begin{table}[t]
    \centering
    \small
    \caption{
    Decay rate $\alpha$
    }
    \label{tab:alpha}
    \small
    \setlength{\tabcolsep}{5pt}{%
    \begin{tabular}{lcc}
        \toprule
        Branch & ACE & FNS-KF  \\
        \midrule
        Spatial branch & 0.984 & 1.522 \\
        Spectral branch & 0.070 & 0.044 \\
        \bottomrule
    \end{tabular}%
    }
\end{table}

We group valid source displacements into unit-width annuli,
\begin{equation}
\begin{gathered}
    A_k=\{\delta:\ q+\delta\in\Omega_h,\;
    1\leq\|\delta\|_2\leq32,\;
    k\leq\|\delta\|_2<k+1\}, \\
    d_b^{(n)}(k)=\frac{1}{|A_k|}
    \sum_{\delta\in A_k}\widetilde K_b^{(n)}(q,q+\delta).
\end{gathered}
    \label{eq:diagnostic_radial}
\end{equation}
Only positions inside the feature grid are included. $d_b^{(n)}(k)$ denotes annular means.

\subsection{Effective decay rate}
The transition product in \eqref{eq:kernel_local} motivates quantifying how effective-kernel strength attenuates with distance. We therefore introduce an effective exponential decay rate $\alpha$ to summarize the measured radial profiles. For each sample $n$ and branch $b$, we fit
\begin{equation}
    \log d_b^{(n)}(r)
    \approx \beta_b^{(n)}-\alpha_b^{(n)}r,
\end{equation}
by ordinary least squares. Larger $\alpha$ indicates faster radial attenuation.

As shown in Table~\ref{tab:alpha}, the spatial branch has a
substantially larger decay rate than the spectral branch on
both datasets, supporting their complementary roles in
localized and domain-wide interactions.
This contrast is more pronounced on FNS-KF, where faster
spatial decay accompanies weaker spectral attenuation,
consistent with the coexistence of local transport and
nonlocal coupling in incompressible flow.
For the spectral branch, $\alpha$ summarizes the overall
attenuation trend rather than a strict exponential decay law.

\subsection{Spatial branch variants}
Figure~\ref{fig:ace-effective-kernel}(c) compares the spatial branches of separately trained one-, four-, and eight-directional variants. A single traversal produces a predominantly one-sided response, four traversals extend influence along both coordinate axes, and eight additionally strengthen diagonal interactions. The comparison shows that additional scan directions broaden spatial coverage, although the kernels remain direction-dependent.

\section{Qualitative Evaluation}
\label{app:qualitative_evaluation}
\begin{figure}[p]
    \centering
    \includegraphics[
        width=\textwidth,
        height=0.78\textheight,
        keepaspectratio
    ]{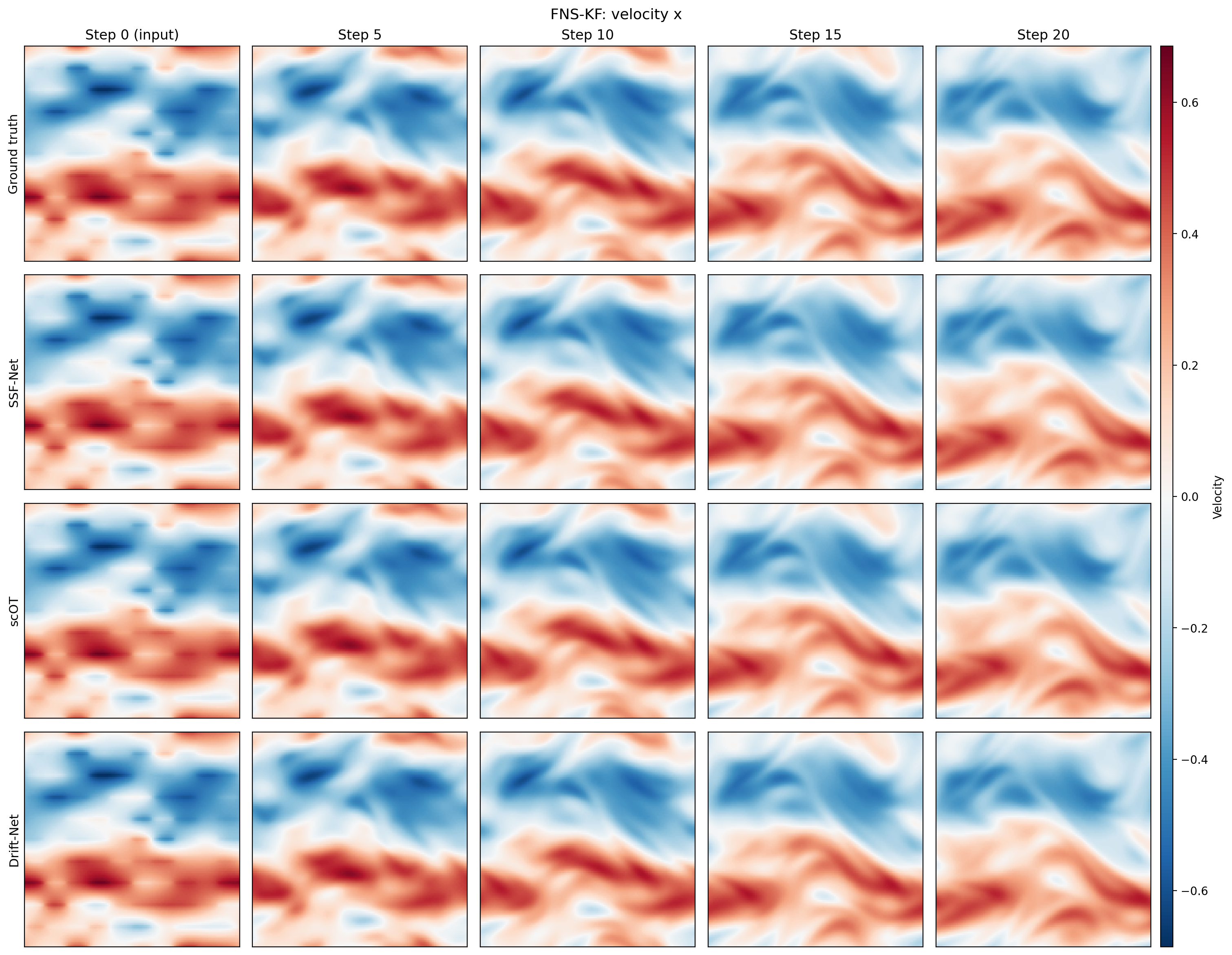}
    \caption{
    \revised{
        Ground-truth and autoregressive predictions of the
        horizontal velocity $u_x$ for a FNS-KF test trajectory.
        Rows show the ground truth, SSF-Net, scOT, and Drift-Net.
        Columns show rollout steps 0, 5, 10, 15, and 20.
        Step 0 is the input shared by all models.
        }
    }
    \label{fig:fns-kf-ux-values}
\end{figure}

\begin{figure}[p]
    \centering
    \includegraphics[
        width=\textwidth,
        height=0.78\textheight,
        keepaspectratio
    ]{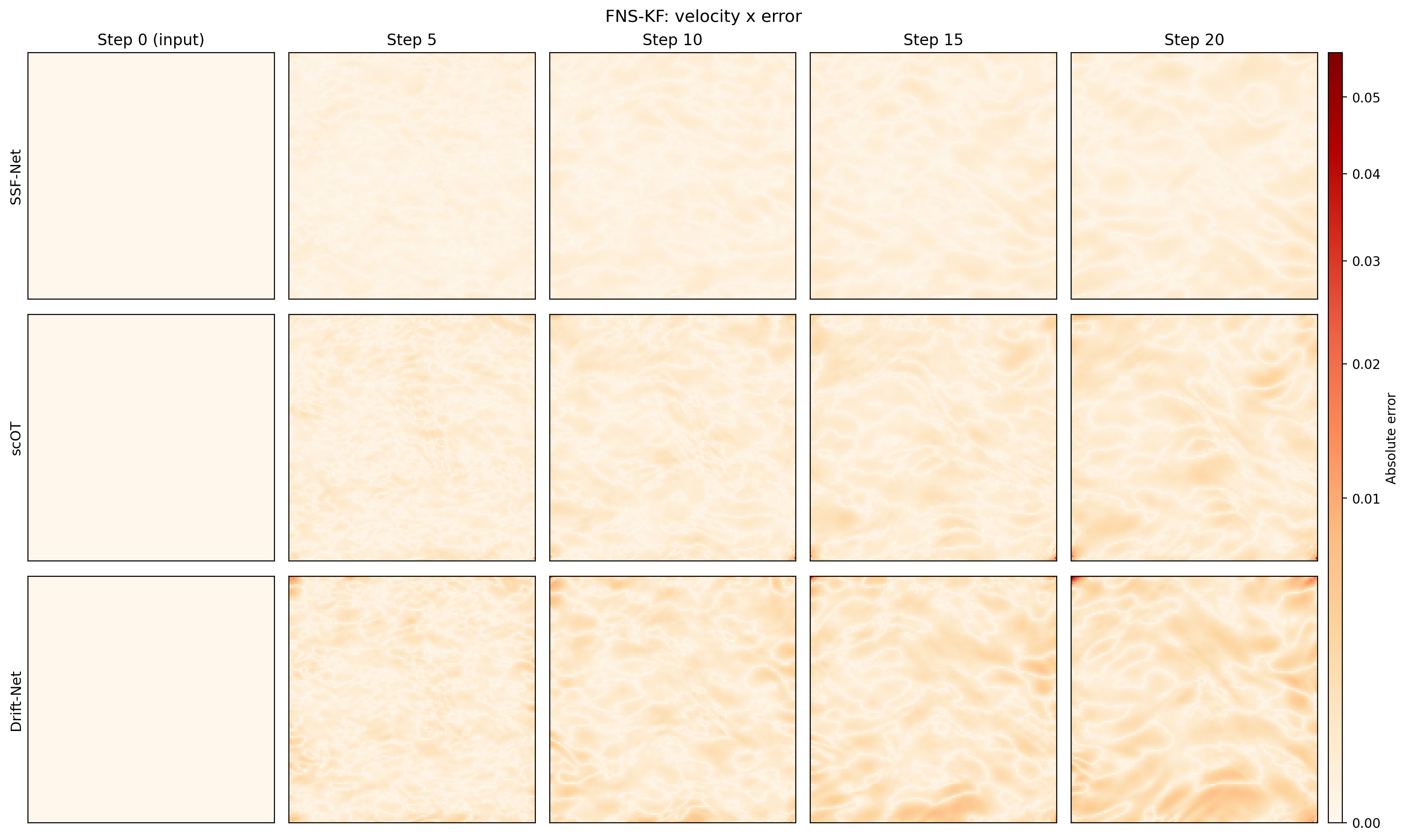}
    \caption{
    \revised{
        Pointwise absolute errors in horizontal velocity $u_x$
        for the trajectory in Figure~\ref{fig:fns-kf-ux-values}.
        Rows show SSF-Net, scOT, and Drift-Net;
        columns show rollout steps 0, 5, 10, 15, and 20.
        }
    }
    \label{fig:fns-kf-ux-error}
\end{figure}

\begin{figure}[p]
    \centering
    \includegraphics[
        width=\textwidth,
        height=0.78\textheight,
        keepaspectratio
    ]{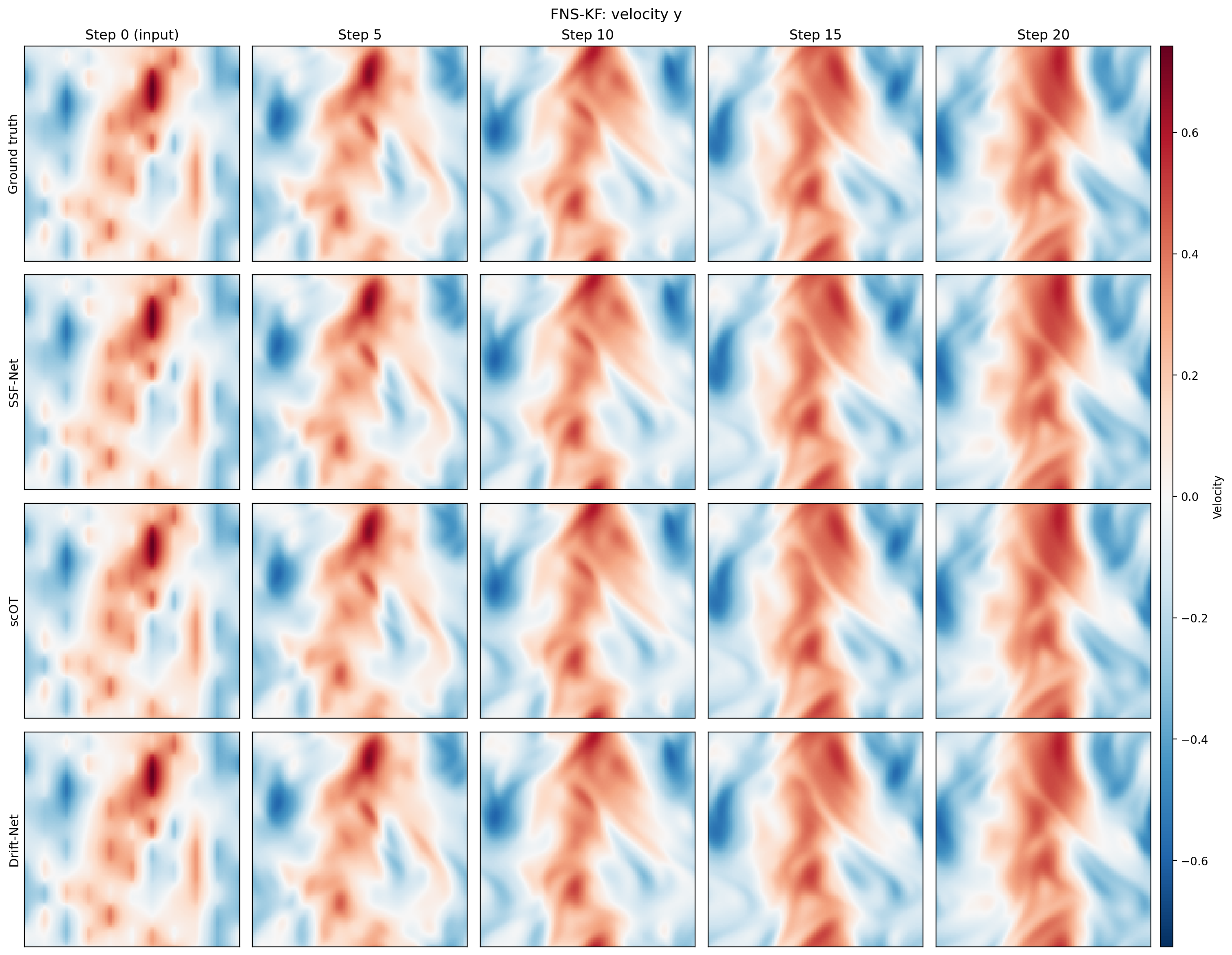}
    \caption{
    \revised{
        Ground-truth and autoregressive predictions of the
        vertical velocity $u_y$ for the same FNS-KF test trajectory.
        Rows show the ground truth, SSF-Net, scOT, and Drift-Net.
        Columns show rollout steps 0, 5, 10, 15, and 20.
        Step 0 is the input shared by all models.
        }
    }
    \label{fig:fns-kf-uy-values}
\end{figure}

\begin{figure}[p]
    \centering
    \includegraphics[
        width=\textwidth,
        height=0.78\textheight,
        keepaspectratio
    ]{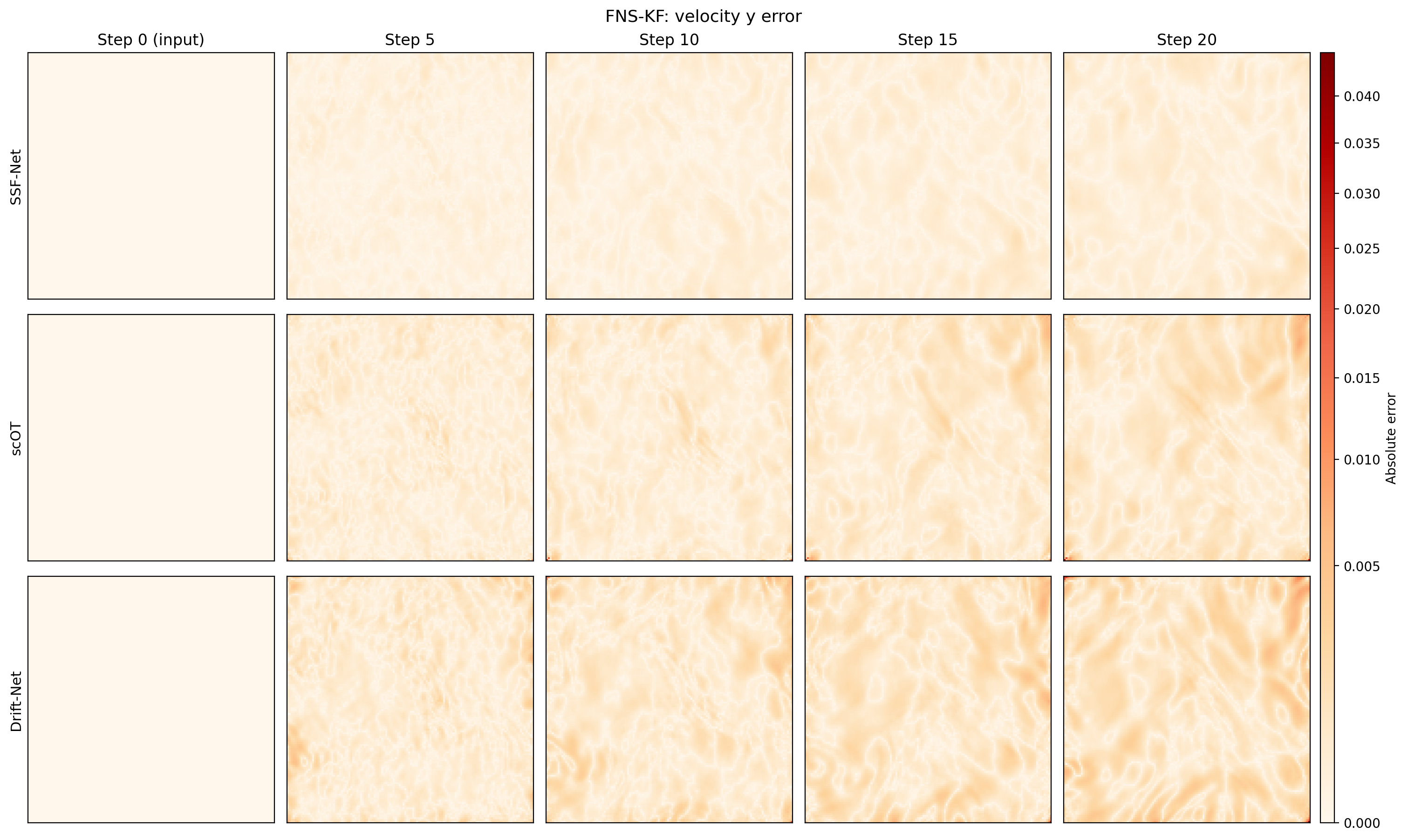}
    \caption{
    \revised{
        Pointwise absolute errors in vertical velocity $u_y$
        for the trajectory in Figure~\ref{fig:fns-kf-uy-values}.
        Rows show SSF-Net, scOT, and Drift-Net;
        columns show rollout steps 0, 5, 10, 15, and 20.
        }
    }
    \label{fig:fns-kf-uy-error}
\end{figure}

We visualize the two velocity components of a FNS-KF test trajectory at rollout steps 0, 5, 10, 15, and 20, comparing SSF-Net, scOT, and Drift-Net against the ground truth. Figures~\ref{fig:fns-kf-ux-values} and~\ref{fig:fns-kf-uy-values} show the predicted fields, while Figures~\ref{fig:fns-kf-ux-error} and~\ref{fig:fns-kf-uy-error} show the corresponding pointwise absolute errors. Step 0 is the supplied initial condition, so its error is zero for all models.

\end{document}